\documentclass[conference]{IEEEtran}
\IEEEoverridecommandlockouts
\usepackage{cite}
\usepackage{enumitem}
\usepackage{multirow}
\usepackage{graphicx}
\usepackage{textcomp}
\usepackage{xcolor}
\usepackage{float}
\usepackage{booktabs}
\usepackage{makecell}
\usepackage{subcaption}
\usepackage{wrapfig}
\usepackage{bbding}
\usepackage{amsmath,amssymb,amsfonts}
\usepackage[ruled,linesnumbered]{algorithm2e}
\usepackage{soul}
\def\BibTeX{{\rm B\kern-.05em{\sc i\kern-.025em b}\kern-.08em
    T\kern-.1667em\lower.7ex\hbox{E}\kern-.125emX}}
\begin{document}

\title{Training-free Heterogeneous Graph Condensation via Data Selection}
\author{
\IEEEauthorblockN{Yuxuan Liang\IEEEauthorrefmark{1},
 Wentao Zhang\IEEEauthorrefmark{2},
Xinyi Gao\IEEEauthorrefmark{3}, 
Ling Yang\IEEEauthorrefmark{4},
Chong Chen\IEEEauthorrefmark{7},
Hongzhi Yin\IEEEauthorrefmark{3},\\
Yunhai Tong\IEEEauthorrefmark{1},
Bin Cui\IEEEauthorrefmark{4}\IEEEauthorrefmark{5}}
\IEEEauthorblockA{
\IEEEauthorrefmark{1}School of Intelligence Science and Technology, Peking University\\
\IEEEauthorrefmark{2}Center for Machine Learning Research, Peking University
\IEEEauthorrefmark{3}The University of Queensland
\\
\IEEEauthorrefmark{4}School of CS \& Key Laboratory of High Confidence Software Technologies (MOE), Peking University\\
\IEEEauthorrefmark{5}Institute of Computational Social Science, Peking University (Qingdao)
\IEEEauthorrefmark{7}Huawei Cloud BU
}
\IEEEauthorblockA{\{liangyx, yangling\}@stu.pku.edu.cn, \{wentao.zhang, yhtong, bin.cui\}@pku.edu.cn, \\
\{xinyi.gao, h.yin1\}@uq.edu.au,
\{chenchong55\}@huawei.com
}
}

\maketitle
\begin{abstract}
Efficient training of large-scale heterogeneous graphs is of paramount importance in real-world applications. However, existing approaches typically explore simplified models to mitigate resource and time overhead, neglecting the crucial aspect of simplifying large-scale heterogeneous graphs from the data-centric perspective. Addressing this gap, HGCond introduces graph condensation (GC) in heterogeneous graphs and generates a small condensed graph for efficient model training. Despite its efficacy in graph generation, HGCond encounters two significant limitations. The first is low effectiveness, HGCond excessively relies on the simplest relay model for the condensation procedure, which restricts the ability to exert powerful Heterogeneous Graph Neural Networks (HGNNs) with flexible condensation ratio and limits the generalization ability. The second is low efficiency, HGCond follows the existing GC methods designed for homogeneous graphs and leverages the sophisticated optimization paradigm, resulting in a time-consuming condensing procedure. In light of these challenges, we present the first Training \underline{Free} \underline{H}eterogeneous \underline{G}raph \underline{C}ondensation method, termed \textbf{FreeHGC}, facilitating both efficient and high-quality generation of heterogeneous condensed graphs. Specifically, we reformulate the heterogeneous graph condensation problem as a data selection issue, offering a new perspective for assessing and condensing representative nodes and edges in the heterogeneous graphs. By leveraging rich meta-paths, we introduce a new, high-quality heterogeneous data selection criterion to select target-type nodes. Furthermore, two training-free condensation strategies for heterogeneous graphs are designed to condense and synthesize other-types nodes effectively.
Extensive experiments demonstrate the effectiveness and efficiency of our proposed method. Besides, FreeHGC exhibits excellent generalization ability across various heterogeneous graph neural networks.
Our codes are available at https://github.com/PKU-DAIR/FreeHGC.

\end{abstract}


\section{Introduction}
Heterogeneous graphs, which contain various types of nodes and edges along with rich semantic information, are prevalent across numerous domains, such as traffic network~\cite{DBLP:conf/icde/HuG0J19,DBLP:conf/kdd/LiHCSWZP19}, biology~\cite{DBLP:conf/sigmod/Vretinaris0EQO21,DBLP:conf/kdd/Do0V19}, and relational databases~\cite{relation_databases1, relation_databases2, relation_databases3}, etc.
As a powerful model for learning from heterogeneous graphs, heterogeneous graph neural networks (HGNNs) have aroused lots of concern in both academia and industry in recent years. They have demonstrated considerable success in various downstream tasks, including node classification, link prediction, and graph clustering~\cite{HGB, HAN, dse1, dse2, scis2, jcst1, jcst2}.

\begin{figure}[t]
\centerline{\includegraphics[width=3.5in]{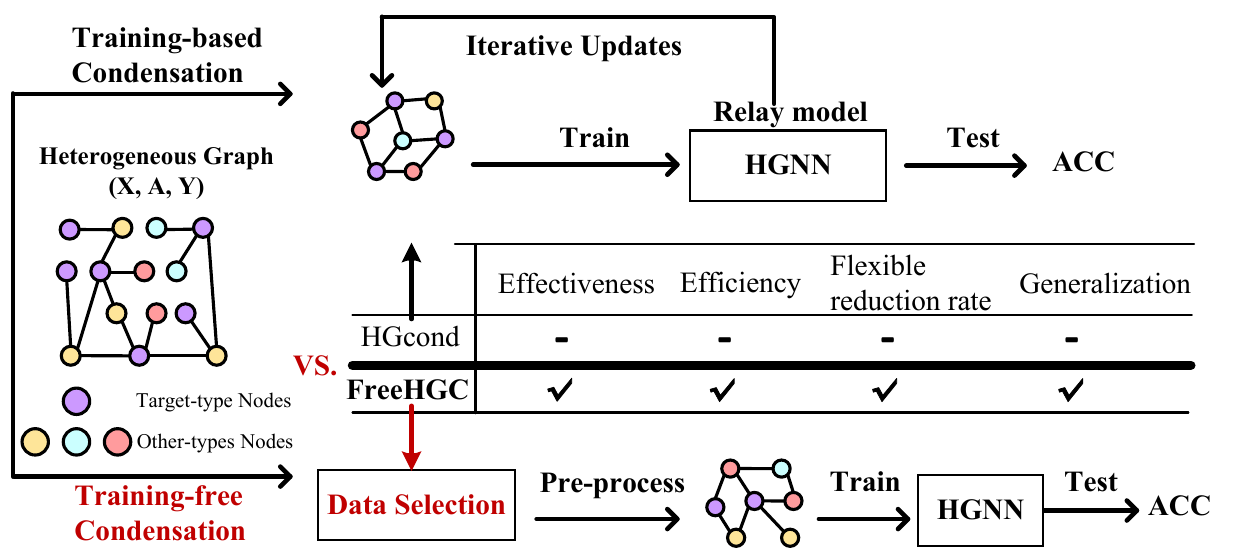}} 
\caption{Comparisons of HGCond and our proposed training free heterogeneous graph condensation method FreeHGC.}
\label{fig:1}
\vspace{-0.5cm}
\end{figure}

Despite the success of HGNNs~\cite{HAN,HGT,HGB,RGCN} on small or medium-scale graphs, many real-world graphs can be very large size in terms of millions of nodes and edges.
Processing such large-scale heterogeneous graphs imposes high computational and storage demands.
To address these issues, two primary research directions have emerged.
The first approach is model-centric, which aims to develop simpler and more efficient model architectures. Representative methods, such as NARS~\cite{Nars} and SeHGNN~\cite{SeHGNN}, move the neighbor aggregation to the pre-processing stage, thereby alleviating the computational burden during online training. But this direction may still not satisfy ever-growing large heterogeneous graphs. Furthermore, the computational cost continues to increase in scenarios that require training multiple models, e.g., multi-stage training~\cite{SAGN, GAMLP}, and hyper-parameter and neural architecture search~\cite{NAS}.
The other approach focuses on simplifying heterogeneous graphs from the data-centric perspective.
The commonly used strategies are heterogeneous graph sparsification~\cite{HSparsification} and heterogeneous graph coarsening~\cite{RSHN}.
These methods either ignore the graph structural information (heterogeneity) or ignore the semantic information (including node features and meta-paths), resulting in sub-optimal HGNN downstream performance.


Recently, graph condensation (GC) has been proposed as a promising solution to intensive computation problem~\cite{Gcond, gao2024graph}.
The graph condensation aims to condense the large original graph through learning a synthetic graph structure and node attributes. 
As a key design, GC utilizes the relay model to connect both the original graph and synthetic (a.k.a condensed) graph, facilitating the comparison of two graphs and the condensation optimization.
Following the gradient matching paradigm of GCond, HGCond~\cite{HGCond} is the first proposed work to condense heterogeneous graphs. Different from GC, it uses clustering information for hyper-node initialization and adopts an orthogonal parameter sequences (OPS) strategy to explore parameters.
Although this method can condense the heterogeneous graph, 
it still suffers from two limitations:

\textbf{Limitation 1: Low effectiveness.} 
Considering the overfitting problem caused by model complexity~\cite{HGCond}, HGCond is forced to use only the simplest heterogeneous graph model as the relay model for graph condensation, it has a large gap between its condensation accuracy and that of state-of-the-art (SOTA) HGNNs~\cite{HAN, HGB, HGT, SeHGNN}.
Even using advanced HGNNs as a relay model yields worse performance.
Besides, the complex optimization problem makes the performance of HGCond degrade or flatten as the size of the condensed graph increases.
Moreover, since the gradient matching loss used by HGCond is model-dependent, the condensed graph may not generalize well to new HGNN models. Specifically, the relay model used by HGCond simply averages different semantics. Due to the distinct semantic fusion methods employed by each heterogeneous graph model, utilizing the condensed heterogeneous graph from HGCond directly across various HGNNs may result in poor generalization.

\textbf{Limitation 2: Low efficiency.} 
Following the paradigm of homogeneous graph condensation method GCond, HGCond requires a bi-level optimization~\cite{Gcond} and nested loop (multiple inner iterations and outer iterations) to jointly learn three objectives: HGNN parameters, condensed node attributes, and heterogeneous graph topology structures. Such complex condensation procedure is computationally intensive and time-consuming. 
Besides, HGCond needs additional clustering information cost for hyper-node initialization and OPS optimization for parameter exploration.
For example, it takes about 1 hour (running on a single TITAN RTX GPU) to condense the large-scale dataset AMiner~\cite{AMiner} to 1\% by 128 epochs.

To address the above two challenges, in this paper, we propose a new Training Free Heterogeneous Graph Condensation method, termed FreeHGC, to select and synthesize high-quality graphs from the original graph structure without model training procedure.
Different from conventional heterogeneous graph condensation that iteratively trains the relay model to optimize the synthetic graph and parameters,
as shown in Figure~\ref{fig:1}, our proposed FreeHGC is model-agnostic and only condenses the graph in the pre-processing stage.
Figure~\ref{fig:1} also highlights the superiority of FreeHGC from four key criteria compared with HGCond: effectiveness, efficiency, flexible condensation ratio, and generalization.
Overall, FreeHGC consists of two key components: (1) a unified data selection metric to select highly informative target-type nodes and (2) two graph condensation strategies for effectively selecting and synthesizing other-types nodes.
To select highly informative target-type nodes, FreeHGC proposes a new heterogeneous graph data selection criterion that unifies the receptive field maximization function and meta-paths similarity minimization function into a unified score objective.
Specifically, FreeHGC introduces the general meta-paths generation model to capture rich semantic information of various meta-paths.
For each meta-path, FreeHGC utilizes the data selection criterion to select target-type nodes with high scores and uses the maximization scores objective to aggregate related meta-paths to obtain high-quality target-type nodes.
Afterward, FreeHGC proposes two condensation strategies to further condense other-types nodes. 
The first strategy constructs the neighbor importance maximization function to filter out unimportant nodes.
Based on the first strategy, the second strategy simulates the aggregation process of nodes to minimize the information loss of synthetic nodes.
Consequently, the condensed target-type nodes and other-types nodes and their connections constitute the condensed graph of FreeHGC.

The main contributions of this paper are as follows:
\begin{itemize}[leftmargin=*]
\item {\textbf{New perspective.}}
We are the first attempt to condense the heterogeneous graph into a small graph without model training. Different from the traditional heterogeneous graph condensation method, we open up a new perspective by transforming the heterogeneous graph condensation problem into the heterogeneous data selection problem.
\item \textbf{{New method.}}
We propose a training-free heterogeneous graph condensation framework named FreeHGC to condense the heterogeneous graph. For condensing target-type nodes, we propose a unified data selection criterion based on the direct and indirect influence of meta-paths over the heterogeneous graph. For condensing other-types nodes, two condensation strategies are designed to select and synthesize nodes for different topology structures.
\item \textbf{SOTA performance.} 
We evaluate the excellent effectiveness and efficiency of FreeHGC on four middle-scale datasets (ACM~\cite{HAN}, DBLP~\cite{HGB}, IMDB~\cite{HGB}, and Freebase~\cite{Freebase}) and one large-scale dataset (Aminer~\cite{metapath2vec}). 
Empirical results also show that FreeHGC outperforms the SOTA heterogeneous graph condensation method in generalization and scalability.
\end{itemize}

\section{Preliminaries}
In this section, we first introduce the notations and problem formulation, then review some existing works related to ours.
\subsection{Notations and Problem Formulation}
A heterogeneous graph (HG) can be defined as $\mathcal{A}=(\mathcal{V}, \mathcal{E}, \phi, \psi)$, where $\mathcal{V}$ is the set of nodes and $\mathcal{E}$ is the set of edges. 
HG contains target-type and other-types nodes. The target-type nodes are used for downstream task prediction.
$\phi(v)$ denotes the node type of each node $v$, $\psi(e)$ denotes the edge type of each edge $e$.
The set of all node types is denoted by $\mathcal{T}= \{\phi(v): \forall v\in \mathcal{V}\}$, the set of all edge types is denoted by $\mathcal{R} = \{\psi(e): \forall e\in \mathcal{E}\}$. 
Each node $v_i \in \mathcal{V}$ has a associated node type object $o_i=\{\phi(v_i): o_i \in \mathcal{T}\}$.
Each edge has a associated relation type $o_{ij} = \{o_i\gets o_j: o_{ij} \in \mathcal{R}\}$, where $\gets$indicates the node type $o_j$ to the target node type $o_i$.
\textbf{Meta-paths} constitute a fundamental concept in heterogeneous graphs, defined as paths comprising composite relationships with multiple edge types, i.e., $\mathcal{P} \triangleq {o_1}{\gets}\cdots{\gets}{o_n}$ (abbreviated as $\mathcal{P} = o_1,...,o_n$).
A graph is considered as homogeneous when $|\mathcal{T}|= |\mathcal{R}| = 1$.

In this paper, we consider a heterogeneous graph dataset $\mathcal{G} = \{\mathcal{A}, \mathbf{X}, \mathbf{Y}\}$, where $\mathbf{X} = \{X_1, \cdots, X_n: n \in \mathcal{T}\}$ represents the feature matrix of each node type and their feature dimensions are usually inconsistent.
$\mathbf{Y} \in \{0, \cdots, C-1\}^N$ denotes the node labels of target type over $C$ classes, $N$ is the number of nodes of the target type.
We aims to select and condense a small graph $\mathcal{G}' = \{\mathcal{A}', \mathbf{X}', \mathbf{Y}'\}$ from the whole graph $\mathcal{G}$, so that the HGNN trained on $\mathcal{G}'$ can maintain comparable performance to the whole graph $\mathcal{G}$.
Given the condensation ratio $r$, all nodes of each node type $N_{type}$ need to be condensed to the condensation budget $\mathcal{B} = rN_{type}$.

\subsection{Heterogeneous Graph Neural Networks}
Existing HGNNs can be broadly classified into two categories: meta-path-based methods and meta-path-free methods.

\noindent\textbf{Meta-path-based methods} use the pre-defined paths that explicitly specify the type of relationships between nodes to guide the message-passing process of each node. 
Specifically, these methods first aggregate neighbor features along the meta-path and then fuse the semantic information generated by different meta-paths to obtain the final node embedding.
Representative methods such as HAN~\cite{HAN} utilizes meta-paths to incorporate both node-level and semantic-level attention during training.
RGCN~\cite{RGCN} is an extension of GCN that handles graphs with multiple relation types. It applies the GCN framework to relational data modeling, and support link prediction and entity classification tasks.
MAGNN~\cite{MAGNN} devises multiple candidate encoder functions for extracting information from meta-path instances.

\noindent\textbf{Meta-path-free methods} do not use predefined meta-paths for message propagation. Instead, they learn to combine the embeddings of node and edge types in the graph.
Representative methods such as RSHN~\cite{RSHN} proposes a unified model that integrates both the graph and its coarsened line graph, allowing for the embedding of both nodes and edges in heterogeneous graphs, without necessitating any prior knowledge such as meta-path.
HGB~\cite{HGB} adopts a multi-layer GAT network as the backbone and incorporates both node features and learnable edge-type embeddings for attention generation. 
HGT~\cite{HGT} combines node and edge types using a transformer model and employs the relative temporal encoding (RTE) technique to model the dynamic dependencies.

Although the above two methods have been widely proposed in HGNNs, they require expensive neighbor aggregation cost during model training, which limits their application on large-scale graphs. To solve this problem, some non-parametric HGNNs use mean aggregator to aggregate the neighbor features in the pre-processing step and combine them in different ways during model training~\cite{Nars, SeHGNN}. Such methods maintain high performance and scalability.

\subsection{Graph Reduction}
Recent years have more works focused on reducing the size of the graph dataset, which is known as graph reduction. It falls into three distinct strategies: graph sparsification, graph coarsening, and graph condensation~\cite{survey-graph-reduction}.

\textbf{Graph Sparsification \& Graph Coarsening.}
Graph sparsification searches for a subset of nodes and edges from the entire graph that maintains a similar level of quality~\cite{Sparsification1, Sparsification2, scis1}.
The common used methods are Herding~\cite{coreset1} and K-center~\cite{coreset2, coreset3}. Herding selects samples closest to the cluster center, while K-Center selects center samples to minimize the largest distance between a sample and its nearest center.
For heterogeneous graphs, the commonly used method like~\cite{HSparsification} is based on random sampling without exploring the heterogeneity of the graph structure, and cannot maintain high accuracy with low data keep ratio.
Graph coarsening aims to preserve a sufficient amount of information, which involves grouping original nodes into super-nodes, and defining their connections~\cite{graph-coarsening}.

\textbf{Graph Condensation.}
Different from graph sparsification and graph coarsening, graph condensation revolves around condensing knowledge from a large-scale graph dataset to construct a much smaller synthetic graph~\cite{Gcond, one-step, SFGC, Eigenbasis-Matching, graph-skeleton, gao1, gao2, gao3}.
By formulating the GC as the bi-level problem, GCond~\cite{Gcond} recently attempted to match the gradients of GNNs and generate an adjacency matrix for the synthetic graph using a trained relay model. This approach achieved a similar performance to the original graph.
Inspired by~\cite{Gcond}, several notable works are highlighted.~\cite{one-step} models the discrete graph structure as a probabilistic model and proposes a one-step gradient matching scheme to reduce computational overhead.~\cite{SFGC} demonstrates the effectiveness of a structure-free graph condensation method. This method condenses a graph into node embedding based on training trajectory matching.
To enhance the generalization of Graph Condensation, GCEM~\cite{Eigenbasis-Matching} proposes eigenbasis matching for spectrum-free graph condensation.

\textbf{Heterogeneous Graph Condensation.}
However, the graph condensation methods mentioned above mainly target homogeneous graphs and are not suitable for heterogeneous graphs due to the lack of additional label information and the requirement for more complex modeling. To address these challenges, HGCond~\cite{HGCond} utilizes clustering information instead of label information for feature initialization and constructs a sparse connection scheme accordingly. Besides, HGCond found that the simple parameter exploration strategy in GCond leads to insufficient optimization on heterogeneous graphs. Therefore, it introduces an exploration strategy based on orthogonal parameter sequences to address the problem.
Another work Graph-Skeleton~\cite{graph-skeleton} also considers heterogeneous graphs, however, it only condenses background nodes and retains all target nodes, which is different from the general graph condensation settings.

\section{observation and insight}
The existing heterogeneous graph condensation method HGCond adopts the traditional gradient matching paradigm like GCond to condense the graph with the simplest heterogeneous graph model. 
This method, however, may lead to some potential limitations, such as low effectiveness and low efficiency. 
In this section, we conduct empirical analysis to explore these two limitations and offer insights into how FreeHGC overcomes them.

\begin{figure}[t]
    \centering
    \subcaptionbox{Low accuracy on ACM and IMDB.}{
    \begin{subfigure}{0.48\linewidth}
        \includegraphics[width=\linewidth]{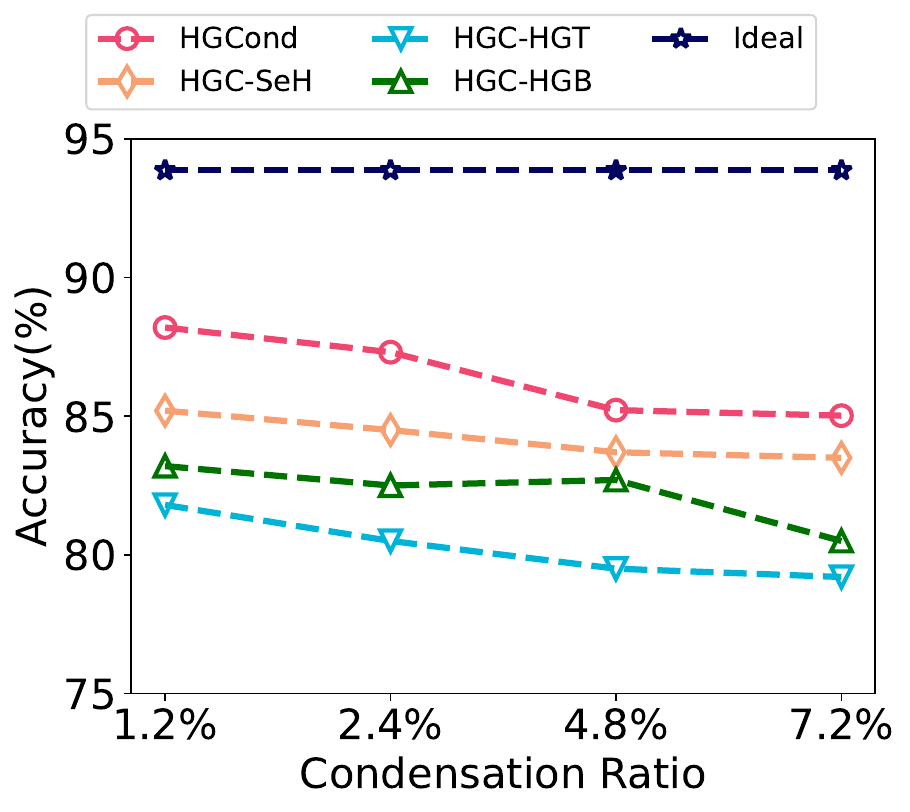}
    \end{subfigure}
    \begin{subfigure}{0.48\linewidth}
        \includegraphics[width=\linewidth]{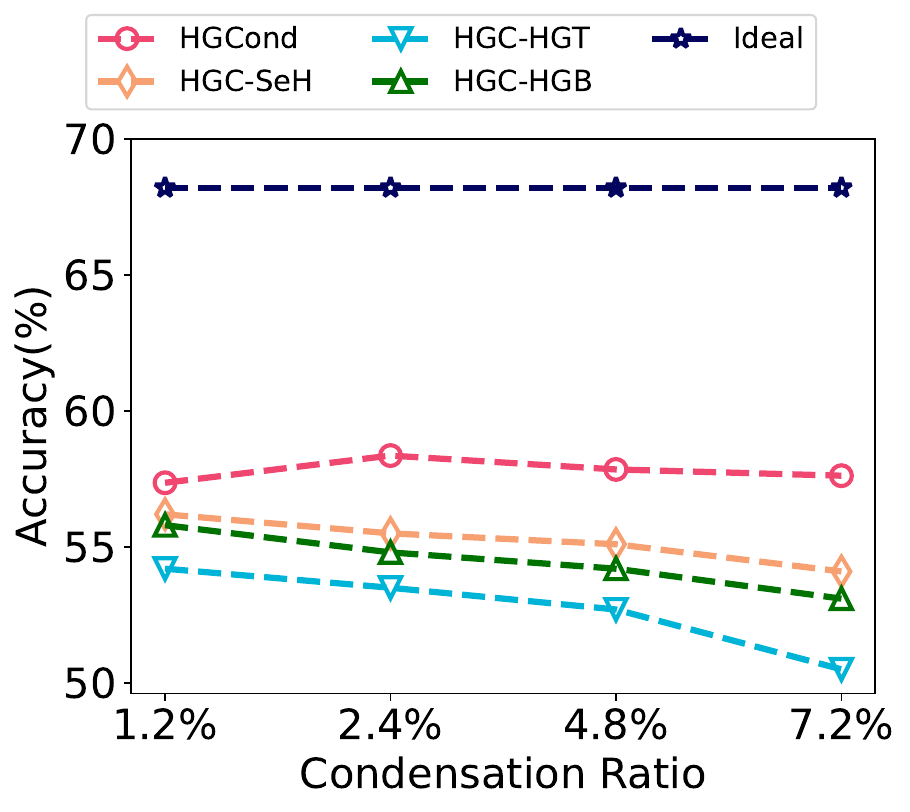}
    \end{subfigure}}
    \subcaptionbox{Low efficiency on Freebase and AMiner.}{
    \begin{subfigure}{0.48\linewidth}
        \includegraphics[width=\linewidth]{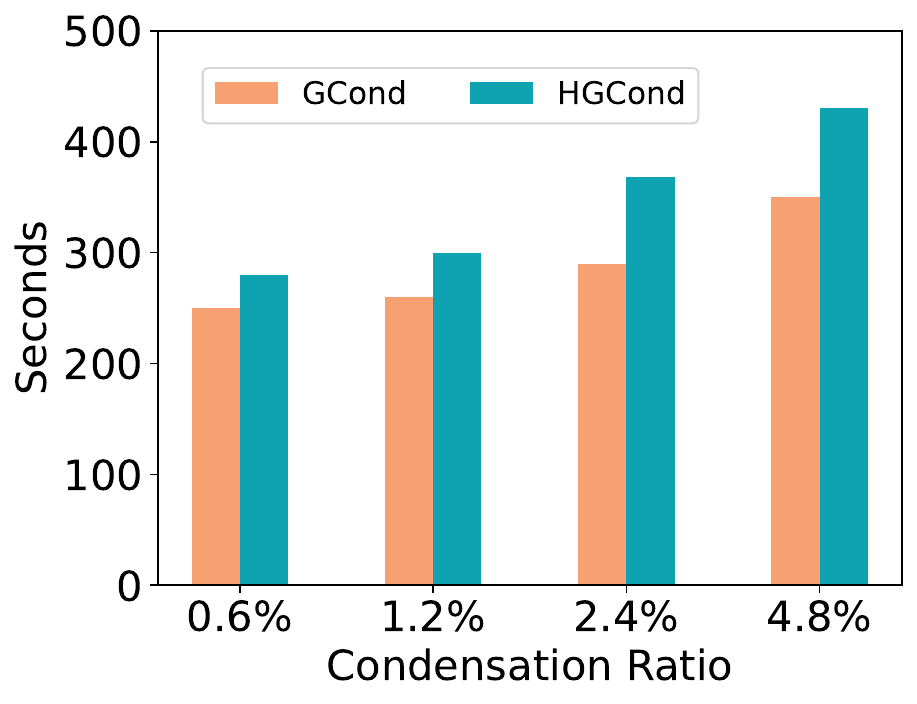}
    \end{subfigure}
    \begin{subfigure}{0.48\linewidth}
        \includegraphics[width=\linewidth]{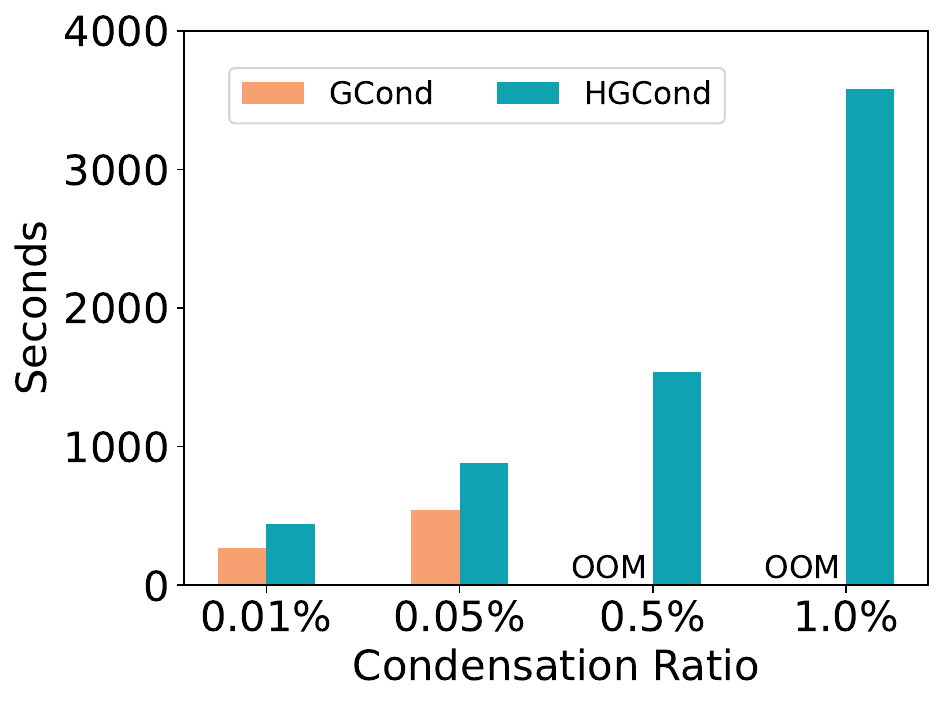}
    \end{subfigure}}
    \caption{Empirical analysis of low accuracy and efficiency.}
\label{fig:motivate}
\vspace{-0.5cm}
\end{figure}

\subsection{Observation on Low Effectiveness}
To ensure the generality of our analysis, we follow HGB benchmark~\cite{HGB} and select two widely used heterogeneous graph datasets: ACM and IMDB. 
In general, low effectiveness can cause two issues: \textbf{low accuracy} and \textbf{poor generalization}.
To reveal the low effectiveness issue, we utilize three representative HGNNs, namely HGT~\cite{HGT}, HGB~\cite{HGB}, and SeHGNN~\cite{SeHGNN} as relay models for HGCond (abbreviated as "HGC-HGT," "HGC-HGB," and "HGC-SeH"), which is much more powerful than the simplest model (HeteroSGC) originally used by HGCond. Among them, SeHGNN is the most powerful SOTA HGNN, and we take its whole-graph prediction accuracy as the ideal result (A: Ideal).
To reveal the poor generalization issue, we use HeteroSGC (abbreviated as HSGC) as the relay model, and evaluate the condensed graph on HSGC, HGT, HGB, and SeHGNN. 
The whole graph accuracy is tested on different HGNNs, abbreviated as “WA”.

\textbf{Low accuracy.} As shown in Figure~\ref{fig:motivate}(a), we can obtain two observations.
(1) The condensation results of HGCond are far from the ideal results (A: Ideal). In the best setting $r$=1.2\%, the accuracy on ACM and IMDB is 87.5 and 57.6, respectively, which can only reach 93.2\% and 84.5\% of the ideal results (A: Ideal) of 93.9 and 68.2.
(2) Using three representative HGNNs (HGT, HGB, and SeHGNN) as relay models for HGCond cannot improve the condensation accuracy. Moreover, as the $r$ ranges from 1.2\% to 7.2\%, the accuracy of all models shows counterintuitive changes, decreasing or flattening instead of increasing.
\textbf{Poor Generalization.}
As shown in Table~\ref{tab:motivate}, we can observe that the performance gap increases when the relay model (HSGC) and evaluation HGNN models have different architectures.
While some studies on GC have addressed this issue~\cite{GCDM, GCEM}, they concentrate on the properties of homogeneous graphs and GNNs, making them inapplicable to HGNNs.

\textbf{Analysis.} (1) The lower accuracy is due to HGCond being parameter-sensitive, as a complex relay model may pose significant challenges to the optimization of GMLoss. Even with OPS-based optimization to improve parameterization, it fails to capture optimal HGNN parameters. 
This limitation further hinders the flexibility of the condensation ratio.
Both GCond and HGCond have mentioned this issue: as the condensation ratio $r$ increases, more hyper-nodes are introduced for a more complex optimization, which raises the risk of overfitting and may converge to local optimality.
(2) The poor generalization stems from the single relay model like HeteroSGC cannot generalize the semantic fusion methods of different HGNNs.

\begin{table}[t]
\renewcommand\arraystretch{1.3}
\caption{\textbf{Poor generalization across different HGNN models, r=2.4\%, SeHGNN is abbreviated as SeH.}}
\centering
\begin{tabular}{c|p{0.5cm}p{0.5cm}|p{0.5cm}p{0.5cm}|p{0.5cm}p{0.5cm}|p{0.5cm}p{0.5cm}}%
\hline
Dataset&HSGC&\pmb{WA}&HGT&\pmb{WA}&HGB&\pmb{WA}&SeH&\pmb{WA}\\
\hline
ACM&86.6&$\pmb{92.5}$&81.4&$\pmb{91.0}$&82.6&$\pmb{93.4}$&87.3&$\pmb{93.9}$\\
IMDB&58.1&$\pmb{58.5}$&55.9&$\pmb{67.2}$&56.4&$\pmb{67.4}$&58.4&$\pmb{68.2}$\\
DBLP&92.8&$\pmb{94.1}$&88.1&$\pmb{93.5}$&89.5&$\pmb{94.5}$&93.0&$\pmb{95.2}$\\
Freebase&53.5&$\pmb{58.3}$&49.3&$\pmb{60.5}$&50.5&$\pmb{66.3}$&54.3&$\pmb{63.4}$\\
\hline
\end{tabular}
\label{tab:motivate}
\vspace{-0.5cm}
\end{table}

\subsection{Observation on Low Efficiency}
To reveal the low efficiency issue, we compare HGCond with GCond (a time-consuming homogeneous graph condensation work), for unlabeled node types, we initialize the hyper-nodes with random sampling from the original nodes to make it applicable to heterogeneous graphs.

As shown in Figure~\ref{fig:motivate}(b), the time cost of HGCond is consistently higher than that of GCond. As the size of the condensed graph grows, the time needed for HGCond increases significantly faster than that of GCond.

\begin{figure*}[t]
\centerline{\includegraphics[width=6.5in]{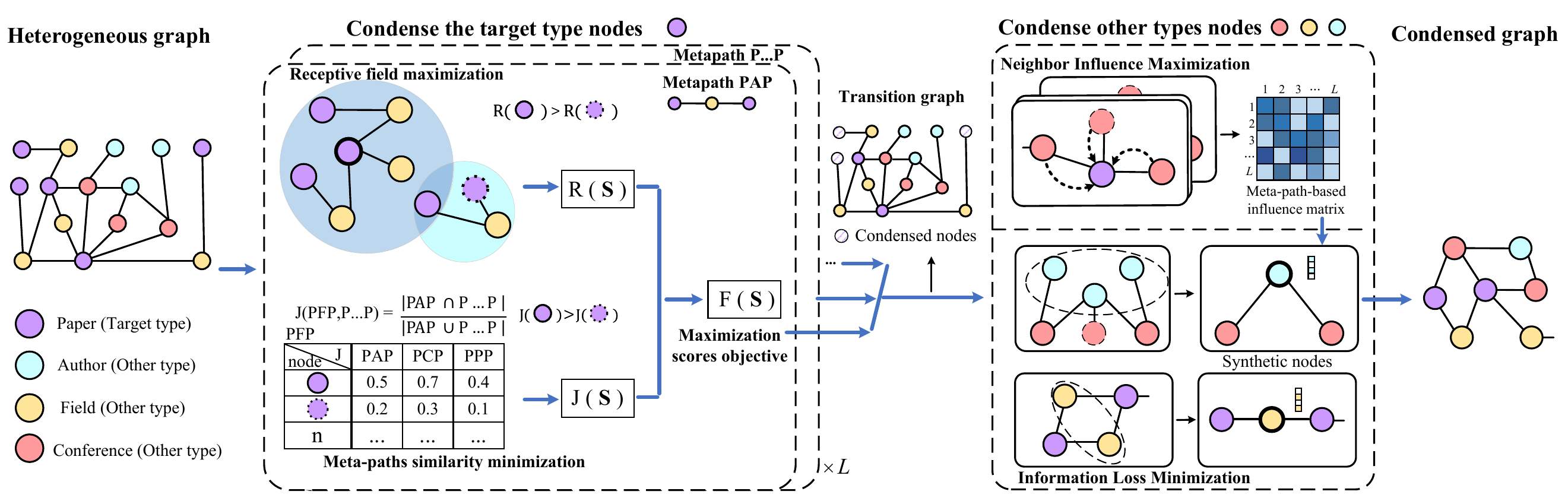}}   
\caption{The workflow of FreeHGC, where $L$ represents the number of meta-paths.}
\label{fig:2}
\vspace{-0.5cm}
\end{figure*}

\textbf{Analysis.}
To ensure the accuracy of the condensed graph, HGCond requires multiple iterations (inner iterations) of updating neural network parameters before updating the synthetic data for multiple iterations (outer iterations). Besides, HGCond needs additional clustering information costs for hyper-nodes initialization and OPS optimization for parameter exploration, which further aggravates the condensation overhead.

\textbf{Our insight.}
In conclusion, the limitation of the constrained relay model makes HGCond unable to take advantage of HGNNs to further improve effectiveness and generalization. 
Given the risk of overfitting, a flexible condensation ratio cannot be employed.
Besides, the limitation of low efficiency makes HGCond take expensive time cost to condense the graph.
Therefore, we expect to employ a simple and powerful heterogeneous graph condensation method without the limitations of relay model.

Our insight is that decoupling condensation from training can avoid the serious optimization challenges brought by complex training, thereby releasing the advantages of HGNN to further improve effectiveness. Once condensation is isolated, the goal of graph condensation is transformed into selecting and condensing high-quality data to protect graph structure information, which has two benefits. First, this allows us to intuitively use a flexible condensation ratio, because the greater the condensation ratio, the better the graph structure information we retain, and the higher the prediction accuracy.
Second, this method does not rely on the semantic aggregation mode of the specified model, so it has good generalization ability for HGNNs.

\section{FreeHGC Framework}
In this section, we propose FreeHGC, the first training-free heterogeneous graph condensation method. As shown in Figure~\ref{fig:2}, our method is divided into two components: condensing the target-type nodes and condensing other-types nodes. The first component uses the receptive field maximization function and meta-paths similarity minimization function, which are based on the direct influence of the graph structure and the indirect influence between meta-paths to calculate the importance of nodes.
Then, FreeHGC combines these two functions as a unified data selection criterion to select high-quality data, which maximizes the influence of nodes while ensuring that each node captures richer graph structural information along different meta-paths.
The second component uses the neighbor importance maximization function to select important father-type nodes and uses the information loss minimization function to synthesize leaf-type nodes.
The above procedure is repeated until the condensed graph is obtained.
Next, we introduce each component of FreeHGC in detail.

\subsection{General Meta-paths Generation Model}
Previous meta-path-based methods rely on pre-defined meta-paths by experts like HAN~\cite{HAN}.
However, this approach requires a lot of manual experience and the selected meta-paths may not be optimal.
Unlike them, we adopt a general method to generate meta-paths inspired by~\cite{SeHGNN}.
Different from~\cite{SeHGNN} to generate aggregated features layer by layer, we separate the feature aggregation operation from meta-path propagation and focus on the graph structure captured by different meta-paths. The graph structure information obtained can help us make full use of the semantic information of heterogeneous graphs for high-quality data selection.
Specifically, We pre-define the maximum hop of meta-paths and utilize all proper meta-paths that are no more than this length. 
Given the target node type $o_t$ and source node type $o_s$, the graph structure information captured by the $k$-hop meta-paths can be expressed as:
\begin{equation}
\begin{aligned}
&\mathbf{\hat{A}}_{o_t,\cdots,o_s} = \mathbf{\hat{A}}_{o_t,o_1}\mathbf{\hat{A}}_{o_1,o_2},\cdots,\mathbf{\hat{A}}_{o_{k-1},o_s},
\end{aligned}
\label{eq-local-multi-2}
\end{equation}
where $\mathbf{{A}}_{o_t,\cdots,o_s}$ is the adjacency matrix of a $k$-hop meta-path, and $\mathbf{\hat{A}}_{o_i,o_j}$ is the row-normalized form of $\mathbf{{A}}_{o_i,o_j}$.

\subsection{Condense the Target-Type Nodes}
Based on the generated meta-paths, we proposes a new high-quality data selection criterion (including two functions mentioned above) to select target-type nodes. To ensure that our data selection criterion have approximate theoretical guarantees, we design the functions that satisfy submodularity~\cite{submodular1}. Here we give the definition of submodular $f$:

\emph{Submodular. Given a node set $S$, the marginal benefit of adding an element to $S$ is at least as high as that of adding the same element to a superset of $S$ (i.e., satisfy the property of diminishing returns): $f(S \cup \{v\}) - f(S) \geq f(W \cup \{v\}) - f(W)$. For all $S \subseteq W \subset V$ and $v \subseteq V \backslash W$.}

Submodular function has the properties of non-decreasing and monotone: $f(S \cup v) \geq f(S)$.
In the following, we take one meta-path to illustrate how our functions work and prove our data selection criterion is submodular.

\underline{Receptive field expansion maximization.}
Unlike images, text, or tabular data, where training data are independently distributed, graph data contains extra relationship information between nodes.
Every node in a $k$-hop propagation will incorporate a set of nodes, including the node itself and its $k$-hop neighbors, which is called Receptive Field (RF).
Since real-world graphs usually have highly skewed power-law degree distributions, the receptive field of each node is inconsistent. As shown in Figure~\ref{fig:2}, we capture the 2-hop neighbors along the $PAP$ meta-path. The receptive fields of target nodes with high degrees are significantly larger than those with low degrees, which means that nodes with large receptive fields can capture more graph structure information.

However, simply finding the maximum receptive field of each node will produce suboptimal results, since the receptive fields may have overlapping parts between nodes, thus limiting the search for the global maximum receptive field for the candidate node set.
To maximize the global receptive field of the current meta-path, we use influence maximization (IM), which aims to select $\mathcal{B}$ nodes so that the number of nodes activated (or influenced) in the social networks is maximized~\cite{SM}:
\begin{equation}
\begin{aligned}
&\mathop{max}\limits_{S} |R(S)|, \mathbf{s.t.} S\subseteq \mathcal{V}, |S| = \mathcal{B},
\end{aligned}
\label{eq-local-multi-2}
\end{equation}
where $R(S)$ represents the set of nodes activated by the candidate node sets $S$ under specific influence propagation models.
Given the selected pool $\mathcal{V}_{train}$, the budget $\mathcal{B}$ and a meta-path from source-type nodes to the target-type nodes $\phi_i \triangleq {o_t}{\gets}\cdots{\gets}{o_s}$. The receptive field maximization function can be defined as:
\begin{equation}
\begin{aligned}
&\mathop{R(S)}\limits_{max}^{\phi_i} = {\bigcup}_{S \in \mathcal{V}_{train}} RF(S).
\end{aligned}
\label{eq3}
\end{equation}

Through Eq.~\ref{eq3}, we use the greedy algorithm~\cite{greedy} to find the node set $S$ that maximizes the global receptive field.
For each node $v_i$ in $\mathcal{V}_{train}$, we calculate its receptive field $RF(S+v_i)$ after adding S. If $RF(S+v_i)>RF(S)$, add $v_i$ to $S$ and remove $v_i$ from $\mathcal{V}_{train}$ then start the calculation of the next node.
This procedure is repeated until the condensation budget $\mathcal{B}$ exhausts.
For efficiency, we can leverage prior works on scalable and parallelizable social influence maximization~\cite{influence-max}. For example, we can use random walks to identify and eliminate uninfluential nodes to greatly decrease the computational workload required for evaluating receptive field expansion.

\begin{figure}[t]
\centerline{\includegraphics[width=3.5in]{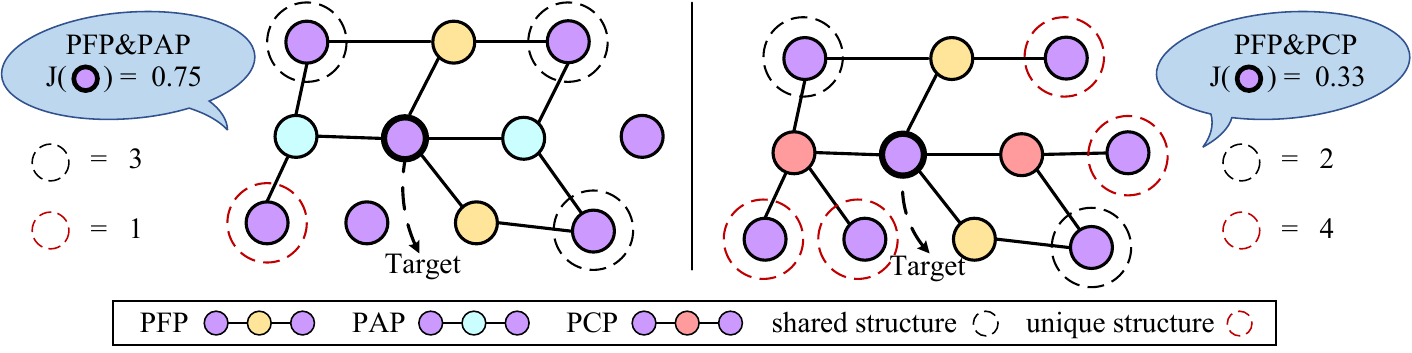}}   
\caption{An illustration of meta-paths similaritiy problem.}
\label{fig:3}
\vspace{-0.5cm}
\end{figure}

A larger $S$ will increase the influence of $R(S)$ and activate more nodes. However, further increasing $S$ reduces the marginal benefit of $R(S)$, proving that $R(S)$ is submodular.
 
\underline{Meta-paths similarity minimization.}
Although the meta-paths can help heterogeneous graphs capture rich graph structure information, there may still be problems in capturing similar heterogeneous graph structure information between different meta-paths.
From the left side of Figure~\ref{fig:3}, we can observe that although the meta-paths $PFP$ and $PAP$ are two different meta-paths. For the target node (purple node in the center), the graph structure information captured by them is almost the same.
In extreme cases, a target node with a high receptive field may capture the same graph structure information along all meta-paths, which means this node can only capture the local graph structure and cannot capture long-distance global information along different meta-paths.

This problem inspired us that simply maximizing $R(S)$ fails to model the interactions between different meta-paths, i.e., the receptive field activated under one meta-path may also be similar under another path, which we refer to as indirect influence between meta-paths.
So besides the direct influence of neighbor information propagation over the graph structure, we also consider strengthening this indirect effect at each node to further improve the diversity of activated nodes $R(S)$ between different meta-paths.
Specifically, we expect the similarity of each node to capture the graph structure information under different meta-paths to be minimal, aiming to maximize the ability of each node to capture different regions of the entire graph.
Here, we introduce the Jaccard index to increase the diversity of nodes.
The Jaccard index~\cite{jaccard} is a well-known measurement of the similarity between two sets $M$ and $T$, defined as the size of the intersection divided by the size of the union of the two sets.
\begin{equation}
\begin{aligned}
J(M, T) = {|M \cap T|}/{|M \cup T|}.
\end{aligned}
\label{eq-local-multi-2}
\end{equation}

We modify it for similarity assessment between different meta-paths. Specifically, given a set of paths $\Phi_L = \{{\phi_i}, i\in L\}$ with the same source type $o_s$ and target type $o_t$, We calculate the Jaccard index for each meta-path. The Jaccard index between two meta-paths can be defined as:
\begin{equation}
\begin{aligned}
J(\phi_i, \phi_j) = {|\phi_i \cap \phi_j|}/{|\phi_i \cup \phi_j|}, \; \; \phi_i, \phi_j \in \Phi_{L}.
\end{aligned}
\label{eq-local-multi-2}
\end{equation}
where we say $J(\phi_i, \phi_j) = 1$ if $|\phi_i \cup \phi_j| = 0$. Then we calculate the sum of the Jaccard index of each meta-path with other related meta-paths, which is used to represent the total similarity of each path with all other meta-paths. The normalized formula can be expressed as:
\begin{equation}
\begin{aligned}
\hat J(\phi_i) = {\sum}_{j=1}^{\Phi_{L}} {J(\phi_i, \phi_j)}/(|\Phi_L| - 1).
\end{aligned}
\label{eq-local-multi-2}
\end{equation}

Then we use Eq.~\ref{eq-sim} to find the node set $S$ that minimizes the meta-paths similarity for each meta-path $\phi_{i}$.
\begin{equation}
\begin{aligned}
&\mathop{J(S)}\limits_{min}^{\phi_i} = \mathop{\hat J(\phi_i)}\limits_{S \in \mathcal{V}_{train}}.
\end{aligned}
\label{eq-sim}
\end{equation}

Therefore, $1-J(S)$ represents the diversity maximization, which allows each node $v \in R(S)$ to maximize the global information captured and reduce the duplication across different meta-paths.
As $S$ increases, $1 - J(S)$ will also increase, as it will cause more nodes to be covered globally across different meta-paths.
However, further increasing $S$ will produce more duplicate nodes, resulting in a decrease in marginal gain. 
Previous studies such as \cite{IoU1} and \cite{IoU2} have also proved that $1-J(M, T)$ is submodular. In summary, $1 - J(S)$ is submodular with respect to $S$.

\underline{Proposed Criterion.}
Based on the above two functions, we propose a new HGNN data selection criterion to consider node influence and diversity simultaneously.
Specifically, FreeHGC adopts a unified maximization scores objective:
\begin{equation}
\begin{aligned}
\mathop{max}\limits_{S}^{\phi_i} \mathop{F(S)} = {\mathop{R(S)}\limits^{\phi_i}}/{|\hat R|} + (1-\mathop{J(S)}\limits^{\phi_i}), \; \; \mathbf{s.t.} \; S\subseteq \mathcal{V}, |S| = \mathcal{B}.
\end{aligned}
\label{eq-local-multi-2}
\end{equation}
where $ |\hat R|$ is the normalization factor, commonly chosen as the total number of source-type nodes. 

Since $R(S)$ and $1-J(S)$ are both submodular, $F(S)$ is also submodular~\cite{submodular2}, we wish to find a k-element set $S^*$ for which $F(S)$ is maximized. This is an NP-hard optimization problem. However, some studies have shown that using greedy algorithm can provide an approximation guarantee of $(1-\frac{1}{e})$ (where $e$ is the base of the natural logarithm)~\cite{greedy}.
Therefore, F(S) follows this principle and ensures that the final selected node set $S$ satisfies $F(S)\geq(1-\frac{1}{e}) \cdot F(S^*)$, providing a $(1-\frac{1}{e})$ approximation.

Then, we aggregate the node scores of all meta-paths, and top-k scores nodes with the highest unified maximization scores $F(S)$ are selected as target-type nodes for condensation, denoted as $\mathcal{N}^{o_t}$:
\begin{equation}
\begin{aligned}
S_{target} = topk({\sum}_{i=1}^{l} {\bigcup}_{i=1}^{l} \mathop{max}\limits_{S}^{\phi_i} F(S)).
\end{aligned}
\label{eq-local-multi-2}
\end{equation}
To ensure that the distribution of classes in the condensed graph is consistent with the distribution of the original graph, we assign the classes to the condensed graph class by class, according to the proportion of classes in the original graph.

Algorithm 1 is the pseudocode of condensing the target-type nodes. In line 1, $M$ is the target type meta-paths generated using the general meta-paths influence model. In line 8, $F_m(S)$ is the score of each node in the current metapath $m$.

\begin{algorithm}[t]
{
\small
\label{algorithm}
\caption{Condense Target-type Nodes}\label{algorithm}
\SetKwData{Left}{left}\SetKwData{This}{this}\SetKwData{Up}{up}
\SetKwInOut{Input}{Input}\SetKwInOut{Output}{Output}
\Input{Heterogeneous graph $\mathcal{G}$, hops for propogation $K$, condensation budget $\mathcal{B}$, number of classes $C$}
\Output{Condensed target type nodes $S_{target}$}
$M = General~Meta\mbox{-}paths(\mathcal{G}, K)$ with Eq. (1)\\
\For{$m \in M$}{
    $J_m(S)=J(m)$ with Eq. (7)\\
    \For{$c \in C$}{
        $R_m^c(S) = R(\mathcal{V}_{train}^{c},\mathcal{B})$ with Eq. (3)\\
        $F_m^c(S) = R_m^c(S) + (1-J_m^c(S))$ with Eq. (8)
    }
    $F_m(S) = \bigcup_{c \in C}F_m^c(S)$\\
}
$S_{target} = topk(\sum \bigcup \mathop{max}\limits_{S} F_m(S))$ with Eq. (9)\\
\textbf{Return:} {$S_{target}$}}
\end{algorithm}

\subsection{Condense Other Types Nodes}
\begin{figure}[t]
\centerline{\includegraphics[width=3.3in]{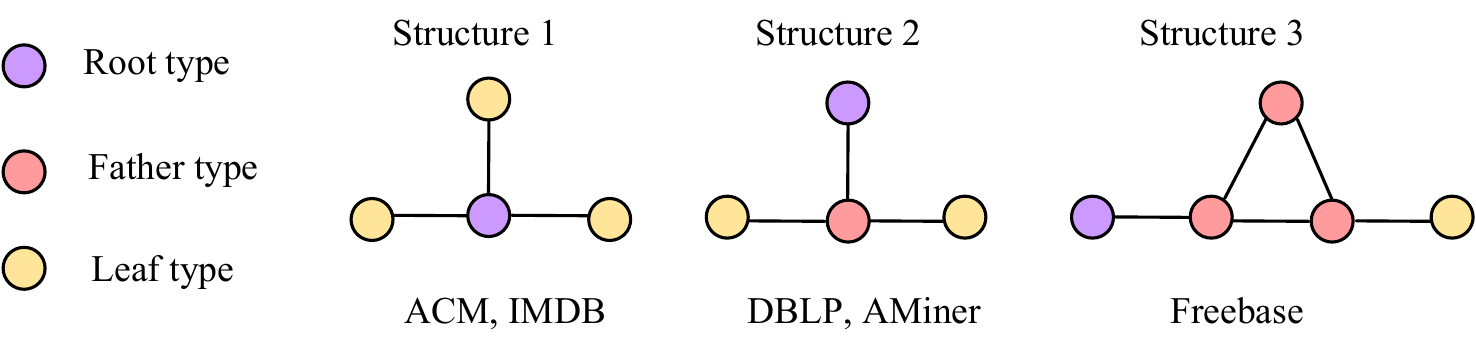}}   
\caption{Topological structures of different datasets.}
\label{fig:top}
\vspace{-0.5cm}
\end{figure}

Unlike homogeneous graph condensation, heterogeneous graph condensation also requires condensation of other node types, which only have node attributes but no labels. 
We classified and summarized the graph structures of commonly used heterogeneous graph datasets (ACM, DBLP, IMDB, Freebase, AMiner) to provide insights for our condensation strategy. As shown in Figure~\ref{fig:top}, we classified different datasets into three topological structures. We then summarized the topological structures into three types from top to bottom according to the vertical hierarchy: root type, father type, and leaf type.
The root type represents the target-type nodes, and the father type and leaf type represent other-types nodes. According to the topological structure, we can see that the father type is a bridge connecting the root type and the leaf type. Directly synthesizing the father type may result in a significant loss of graph structure information between the leaf type and the father type. Based on this insight, we introduce two condensation strategies: the neighbor influence maximization function for the father type and the information loss minimization function for the leaf type.

\underline{Neighbor Influence Maximization.}
The goal is to select the most important neighbor nodes (father type) to be connected to the target nodes (root type). 
Specifically, we measure the importance of other-types nodes to the target-type nodes in a type-by-type manner and adopt the general meta-paths generation model to capture all relevant semantics without manual definition.
Given the target node type $o_t$, the source node type $o_s$ and the hop length $k$, generate all $L$ meta-paths within $k$ hops $\Phi_L = \{{\phi_i}, i\in L\}$, each $\phi_i$ represents a meta-path ${o_t}{\gets}\cdots{\gets}{o_s}$ from the source type $o_s$ to the target type $o_t$, and its adjacency matrix is represented by $A_{\phi_i}$. 
Here we introduce the neighbor influence maximization function $NIM(\cdot)$ to calculate the importance of neighbor nodes:
\begin{equation}
\begin{aligned}
N^i= NIM(\hat{A}^{sym}_{\phi_i}), \; \; i \in L.
\end{aligned}
\label{eq-local-multi-2}
\end{equation}
where $A_{\phi_i}$ is the symmetric matrix and normalize it with $\hat{A}^{sym}_{\phi_i}$, $N^i$ is the influence matrix for metapath $\phi_i$.
We choose Personalized PageRank because it can calculate the correlation of all neighbor nodes with respect to the target vertex, and can be efficiently computed using the approximation techniques to facilitate the scalability for large-scale HINs~\cite{prfast}:
\begin{equation}
\begin{aligned}
N^i= \alpha(I-(1-\alpha)\hat{A}^{sym}_{\phi_i})^{-1}.
\end{aligned}
\label{eq-local-multi-2}
\end{equation}

Note that NIM can be replaced by other node importance evaluation algorithms like degree betweenness and closeness centrality~\cite{centrality}, hubs and authorities~\cite{authorities}.
Since we adopt the general meta-paths generation model, the information of each neighbor in different meta-paths can be comprehensively captured. We aggregate all captured information to generate the influence matrix:
\begin{equation}
\begin{aligned}
N^s = {\sum}_{i=1}^{L}N^i, \; \; i \in L.
\end{aligned}
\label{eq-local-multi-2}
\end{equation}
where $N^s \in \mathbb{R}^{|o_t|\times |o_s|}$. Then, total influence of each node of father-type can be expressed as:
\begin{equation}
\begin{aligned}
S_{father} = topk(\sum_{i=1}^{\mathcal{N}^{o_t}}N^s_{i,:}),
\end{aligned}
\label{eq-local-multi-2}
\end{equation}
where $\mathcal{N}^{o_t}$ is the neighbor nodes of the target-type. Top-k neighbor nodes with the highest influence scores are selected as the condensed nodes for type $o_s$. We repeat this procedure until all father types are condensed.

\begin{figure}[t]
\centerline{\includegraphics[width=3.3in]{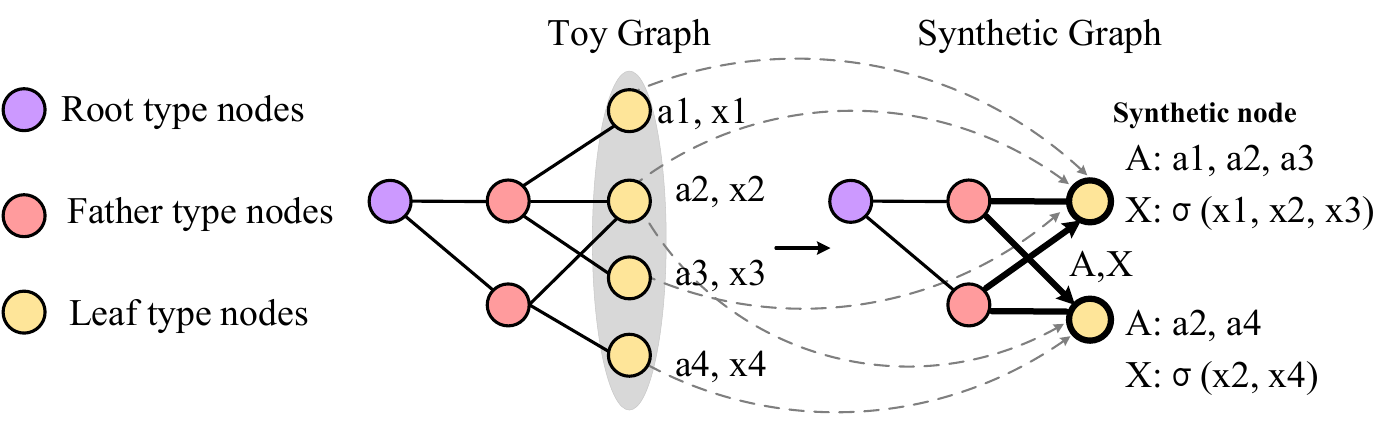}}
\caption{An illustration of information loss minimization.}
\label{fig:cond2}
\vspace{-0.5cm}
\end{figure}


\underline{Information Loss Minimization.}
For leaf-type condensation, our goal is to generate synthetic nodes with minimal information loss (including graph structure and node features). To achieve this goal, we revisit the essence of HGNNs which is neighbor attention of nodes and semantic attention of relations on heterogeneous graphs. The attention methods vary according to different HGNNs.
To study the commonality of attention methods, some work like SeHGNN~\cite{SeHGNN} conducted experiments on different HGNNs. SeHGNN found that replacing the attention node aggregation of HGNNs (such as HAN~\cite{HAN} and HGB~\cite{HGB}) with a mean aggregator does not affect performance, thus concluding that semantic attention is essential while neighbor attention is not necessary.

Based on this finding, we propose a generalized graph condensation method that applies to different HGNNs. The core is to simulate the mean aggregation process of nodes to minimize the information loss of synthetic nodes.
Specifically, for each father-type node $i$, we collect all its neighbor nodes (belonging to leaf-type) $N_i$, and the node feature of each neighbor node $j$ is $X_j, j \in N_i$.
The synthesized hyper-node k is represented as:
\begin{equation}
\begin{aligned}
\mathcal{H}_k= \{e: (i,k), x: \sigma(X_i, i\in N_i)\}
\end{aligned}
\label{XXX}
\end{equation}
where $e$ represents the connected edge, $\sigma(\cdot)$ represents the feature aggregation function, usually using mean aggregator.
As shown in Figure~\ref{fig:cond2}, the toy graph contains one root type, one father type, and one leaf type. Each father-type node identifies and uses the $\sigma()$ function to aggregate its neighbor nodes to form hyper-nodes for leaf-type.
By simulating the mean aggregation process of neighbor nodes, father-type can minimize information loss when aggregating 1-hop leaf-type nodes. However, this method ignores the connection between father-types. In Figure~\ref{fig:cond2}, the father-type nodes can obtain 2-hop information through the leaf type $a2$. Using Eq. 14 will make the father-type node unable to obtain 2-hop information from another connected father-type node.
To solve this problem, we maintain the connection relationship between father-types by adding reverse edges between synthetic nodes and related father-type nodes:
\begin{equation}
\begin{aligned}
{\mathcal{H}_k}^{new}= \mathcal{H}_k \cup \{e: (k\gets j | j\in \mathcal{N}_k, j \neq i)\}
\end{aligned}
\label{XXX}
\end{equation}
where $\mathcal{N}_k$ is the neighbor nodes (father-type) of synthetic node $k$.
For each leaf-type, we repeat the above process until the condensed synthetic node reaches the condensation budget $\mathcal{B}$:
\begin{equation}
\begin{aligned}
S_{leaf}=\cup_{k=1}^{\mathcal{B}}\mathcal{H}_k^{new}
\end{aligned}
\label{XXX}
\end{equation}

For synthetic nodes with lower degrees, we prioritize further condensation.
We repeat this procedure until all leaf-types are condensed.
Our synthesis method not only protects the graph structure information and features but also can be widely used in different HGNN models.

Algorithm 2 is the pseudocode of condensing the other-types nodes. 
In lines 2-4, $S_n$ represents the condensed nodes of father-type $n$, while in line 5, $S_{father}$ represents all condensed father-type nodes. In lines 7-9, we use Eq. 16 to condense leaf-types nodes. In line 11, the condensed nodes $S_{target}$, $S_{father}$, and $S_{leaf}$ obtained from Algorithm 1 and Algorithm 2 collectively constitute the final condensed graph $\mathcal{G}'$.

\begin{algorithm}[t]
{
\small
\label{algorithm}
\caption{Condense Other-types Nodes}\label{algorithm}
\SetKwData{Left}{left}\SetKwData{This}{this}\SetKwData{Up}{up}
\SetKwInOut{Input}{Input}\SetKwInOut{Output}{Output}
\Input{Meta-paths generated by Algorithm 1: $M$, condensed target-type nodes $S_{target}$}
\Output{Condensed heterogeneous graph $\mathcal{G}'$}
\%Condense father-types nodes\\
\For{$n \in father\mbox{-}types$}{
    $S_n = topk(NIM(M_n))$ with Eq. (13)
    }
$S_{father} = \bigcup_{n\in father\mbox{-}types}S_n$\\
\%Condense leaf-types nodes\\
\For{$k \in leaf\mbox{-}types$}{
    get $S_k$ with Eq. (16)
    }
$S_{leaf} = \bigcup_{n\in leaf\mbox{-}types}S_k$\\
$\mathcal{G}' = S_{target} \cup S_{father} \cup S_{leaf}$\\
\textbf{Return:} {$\mathcal{G}'$}}
\end{algorithm}

\textbf{Time Complexity.}
We analyze the time complexity of traditional graph condensation methods GCond and HGCond and compare them with FreeHGC.
Let the number of GCN layers be $L$, the large-scale graph node number be $N$, the small-scale condensed graph node number be $N'$, and the feature dimension be $d$.
The complexity of Gcond consists of two parts: graph condensation and training the entire graph. 
The overall time complexity of GCond~\cite{Gcond} is $TK\mathcal{O}(LN^{'2}d+N^{'2}d^2+LEd+LNd^2)$. Where $T$ denotes the number of iterations (nested loops), $K$ denotes the different initialization.
Although $c$ rounds of iterations are required to
generate synthetic nodes for different classes, it can be easily parallelizable.

The graph condensation process of HGCond is similar to that of GCond. In addition, HGCond requires a clustering algorithm and OPS optimization to generate synthetic nodes for different node types.
The overall time complexity of HGCond~\cite{HGCond} is about $\mathcal{O}(GCond)+\mathcal{O}(\sum_{all~node~types}((\alpha N_{type}-1)N_{(type)}+d_{out, (type)}^3))$. Where $\alpha$ is the condensation ratio, $d_{out, (type)}$ is the output channels.

The complexity of FreeHGC mainly consists of two parts: (1) condense the target types and (2) condense other types. Specifically, the time complexity of (1) is $\mathcal{O}(\alpha N_{tgt}^2 + N_{tgt}logN_{tgt}))$, where $N_{tgt}$ is the number of nodes of target type, $N_{tgt}logN_{tgt}$ is for sorting and finding the top-k scored nodes. 
The classes and meta-paths loop in Algorithm 1 line 2-9 can be easily parallelizable.
The time complexity of (2) needs to condense father-types $types_{f}$ and leaf-types $types_{l}$, where $types_{f} + types_{l} = types_{others}$. 
For neighbor influence maximization, FreeHGC requires $Ktypes_{f}\mathcal{O}( K\frac{E}{\varepsilon})$ to perform the personalized PageRank algorithm, where $E$ is the number of edges corresponding to each meta-path, $\varepsilon$ denote the error threshold. $K$ meta-paths can be simply parallelized, and the time complexity is reduced to $types_{f} \mathcal{O}(\frac{E}{\varepsilon})$.
For influence loss minimization, the time complexity is $types_{l}\mathcal{O}{(\alpha N_{l}d)}$.
Therefore, the overall time complexity of FreeHGC is $\mathcal{O}(\alpha N_{tgt}^2 + N_{tgt}logN_{tgt})+types_{f} \mathcal{O}{(\frac{E}{\varepsilon})}+types_{l}\mathcal{O}{(\alpha N_{leaf}d)}$.
Generally, the time complexity of FreeHGC is significantly lower than
GCond and HGCond due to the training-free paradigm.
\begin{table}[t]
\renewcommand\arraystretch{1}
\caption{\textbf{Overview of the datasets.}}
\setlength{\tabcolsep}{1mm}
\centering
\resizebox{0.45\textwidth}{!}{
\begin{tabular}{*{7}{c}}%
\hline%
\makecell[c]{Datasets}&\#Nodes&\makecell[c]{\#Nodes \\types}&\#Edges&\makecell[c]{\#Edge \\types}&Target&\#Classes\\
\hline%
DBLP&26,128&4&239,566&6&author&4\\
ACM&10,942&4&547,872&8&paper&3\\
IMDB&21,420&4&86,642&6&movie&5\\
Freebase&180,098&8&1,057,688&36&book&7\\
AMiner&4,891,819&3&12,518,010&2&author&8\\
MUTAG&27,163&7&148,100&46&d&2\\
AM&881,680&7&5,668,682&96&proxy&11\\
\hline
\end{tabular}}
\label{tab:Datasets}
\vspace{-0.5cm}
\end{table}

\begin{table*}[t]
\renewcommand\arraystretch{1.2}
\caption{\textbf{Experiment results of node classification prediction tasks on four datasets.}}
\centering
\resizebox{0.9\textwidth}{!}{
\begin{tabular}{*{9}{c}}%
\toprule%
& &\multicolumn{5}{c}{Baselines}& Proposed\\
\cmidrule(r){3-7} \cmidrule(r){8-8}
Dataset&Ratio (r)&Random-HG&Herding-HG&K-Center-HG&Coarsening-HG&HGCond&\pmb{FreeHGC}&Whole Dataset\\
\hline
\multirow{4}{*}{ACM}
&
1.2\%&$53.37\pm0.24$&$63.71\pm0.53$&$62.66\pm0.26$&$64.17\pm0.31$&$88.26\pm6.85$&$\pmb{90.02\pm0.44}$&\multirow{4}{*}{$93.87\pm0.50$}\\
&
2.4\%&$61.18\pm0.36$&$67.66\pm0.63$&$65.45\pm1.21$&$65.56\pm0.24$&$87.31\pm5.31$&$\pmb{91.27\pm0.76}$&\\
&
4.8\%&$60.01\pm2.36$&$72.14\pm0.21$&$69.68\pm0.76$&$68.91\pm0.73$&$85.22\pm6.28$&$\pmb{92.71\pm0.31}$&\\
&
9.6\%&$66.25\pm1.18$&$79.14\pm0.21$&$75.68\pm0.76$&$70.91\pm0.73$&$85.02\pm4.36$&$\pmb{93.62\pm0.31}$&\\
\hline
\multirow{3}{*}{DBLP}
&
1.2\%&$38.73\pm0.98$&$60.37\pm0.51$&$61.39\pm0.26$&$53.27\pm0.22$&$\pmb{93.18\pm0.94}$&${91.35\pm0.54}$&\multirow{4}{*}{$95.24\pm0.13$}\\
&
2.4\%&$48.84\pm2.36$&$65.72\pm1.06$&$63.80\pm0.78$&$59.63\pm0.65$&${93.01\pm0.51}$&$\pmb{93.21\pm0.55}$&\\
&
4.8\%&$45.49\pm0.52$&$72.17\pm0.75$&$70.68\pm0.37$&$66.21\pm0.25$&${92.76\pm0.44}$&$\pmb{94.05\pm0.41}$&\\
&
9.6\%&$56.01\pm2.36$&$80.14\pm0.21$&$79.68\pm0.76$&$76.91\pm0.73$&$92.53\pm0.88$&$\pmb{94.59\pm0.31}$&\\
\hline
\multirow{4}{*}{IMDB}
&
1.2\%&$39.22\pm0.90$&$45.69\pm0.28$&$43.75\pm0.53$&$40.58\pm0.98$&$\pmb{57.36\pm0.44}$&${56.53\pm0.74}$&\multirow{4}{*}{$68.21\pm0.32$}\\
&
2.4\%&$42.65\pm0.73$&$48.85\pm1.13$&$49.71\pm0.13$&$45.66\pm1.25$&${58.36\pm0.44}$&$\pmb{60.75\pm0.82}$&\\
&
4.8\%&$48.28\pm0.58$&$51.69\pm1.32$&$53.88\pm0.26$&$48.85\pm0.31$&${57.85\pm0.44}$&$\pmb{63.23\pm0.31}$&\\
&
9.6\%&$51.01\pm2.36$&$58.14\pm0.21$&$58.60\pm0.76$&$50.91\pm0.73$&$57.62\pm0.31$&$\pmb{66.35\pm0.31}$&\\
\hline
\multirow{4}{*}{Freebase}
&
1.2\%&$45.32\pm0.47$&$49.11\pm0.13$&$48.18\pm0.26$&$46.28\pm0.56$&${53.29\pm0.44}$&$\pmb{54.46\pm0.74}$&\multirow{4}{*}{$63.41\pm0.47$}\\
&
2.4\%&$47.52\pm0.14$&$51.17\pm0.46$&$48.85\pm0.71$&$49.10\pm0.56$&${54.34\pm0.44}$&$\pmb{60.55\pm0.20}$&\\
&
4.8\%&$48.15\pm1.21$&$52.02\pm2.05$&$51.33\pm0.86$&$50.25\pm0.56$&${53.81\pm0.44}$&$\pmb{61.23\pm0.56}$&\\
&
9.6\%&$50.01\pm2.36$&$53.14\pm0.21$&$52.68\pm0.76$&$52.91\pm0.73$&$53.22\pm0.31$&$\pmb{62.15\pm0.31}$&\\
\hline
\end{tabular}}
\label{exp1}
\end{table*}

\begin{table*}[t]
\renewcommand\arraystretch{1.2}
\caption{\textbf{Generalization ability across different HGNN models.}}
\centering
\resizebox{0.8\textwidth}{!}{
\begin{tabular}{*{8}{c}}%
\toprule%
&Methods&HGB&HGT&HAN&SeHGNN&Condensed Avg.& Whole Avg.\\
\hline
\multirow{3}{*}{\makecell{ACM\\r = 2.4\%}}
&
Herding-HG&$66.85\pm0.51$&$65.21\pm0.36$&$63.98\pm0.77$&$67.66\pm0.63$&65.92&\multirow{3}{*}{92.25}\\
&
HGCond&$85.71\pm0.38$&$82.63\pm0.18$&$81.35\pm0.40$&$87.31\pm0.76$&84.25&\\
&
FreeHGC&$90.86\pm0.38$&$89.31\pm0.18$&$88.35\pm0.40$&$91.27\pm0.76$&\pmb{89.95}&\\
\hline
\multirow{3}{*}{\makecell{DBLP\\r = 2.4\%}}
&
Herding-HG&$67.80\pm0.66$&$65.95\pm0.41$&$63.17\pm0.85$&$65.72\pm1.06$&65.66&\multirow{3}{*}{93.81}\\
&
HGCond&$89.45\pm0.81$&$89.07\pm0.36$&$86.16\pm0.26$&$93.01\pm0.18$&89.42&\\
&
FreeHGC&$92.20\pm0.38$&$91.31\pm0.18$&$90.35\pm0.40$&$93.21\pm0.55$&\pmb{91.77}&\\
\hline
\multirow{3}{*}{\makecell{IMDB\\r = 2.4\%}}
&
Herding-HG&$47.28\pm0.26$&$45.79\pm0.46$&$44.80\pm0.56$&$48.85\pm1.13$&47.91&\multirow{3}{*}{66.85}\\
&
HGCond&$56.42\pm0.17$&$55.88\pm0.50$&$54.36\pm1.22$&$58.36\pm0.44$&56.26&\\
&
FreeHGC&$61.52\pm0.38$&$61.31\pm0.18$&$59.86\pm0.40$&$60.75\pm0.82$&\pmb{60.86}&\\
\hline
\multirow{3}{*}{\makecell{Freebase\\r = 2.4\%}}
&
Herding-HG&$49.81\pm0.43$&$47.22\pm0.97$&$43.44\pm0.35$&$51.17\pm0.46$&40.29&\multirow{3}{*}{61.25}\\
&
HGCond&$50.52\pm0.52$&$49.34\pm1.35$&$48.25\pm0.98$&$54.34\pm0.44$&$50.61$&\\
&
FreeHGC&$61.31\pm0.38$&$59.06\pm0.18$&$53.35\pm0.40$&$60.55\pm0.20$&\pmb{58.57}&\\
\hline
\end{tabular}}
\label{exp3}
\end{table*}



\section{Experiments}
In this section, we execute comprehensive experiments to evaluate the proposed FreeHGC for the node classification task in different heterogeneous graph datasets. Specifically, we first introduce experimental settings.
Next, we showcase the advantages of FreeHGC through four distinct perspectives: (1) a comprehensive comparison with representative baselines for effectiveness; (2) a comparison of large-scale dataset (AMiner) for scalability; (3) a comparison with different HGNN models for generalizability; (4) an analysis of condensed data; (5) an ablation experiments for FreeHGC.

\subsection{Datasets and Baselines}
\textbf{Datasets}. Four middle-scale datasets from the HGB benchmark are used, namely ACM, DBLP, IMDB, and Freebase~\cite{HGB}. 
We follow the HGB benchmark that splits the node labels into 24\%/6\%/70\% for training, validation, and testing.
Two knowledge graphs from DGL~\cite{dgl} are used, namely MUTAG and AM.
One large-scale dataset AMiner~\cite{AMiner} from~\cite{metapath2vec}. 
Table~\ref{tab:Datasets} shows the graph structure properties of all datasets.
\begin{itemize}[leftmargin=*]
\item \textbf{Middle-scale datasets}. ACM and DBLP are academic networks. IMDB is a website that contains movies and related media. Freebase is a large knowledge graph. MUTAG and AM are two knowledge graphs with more relations~\cite{RGCN}.
\item \textbf{Large-scale dataset.} 
AMiner was first proposed in~\cite{AMiner}. Then~\cite{metapath2vec} constructs the heterogeneous collaboration network, which consists of three types of nodes: computer scientists, papers, and papers from computer science venues (both conferences and journals) up to 2016.
\end{itemize}

\textbf{Baselines}. To evaluate the performances of FreeHGC, we choose five baselines: three corset methods (Random, Herding~\cite{coreset1} and K-center~\cite{coreset2,coreset3}), one graph coarsening method~\cite{graph-coarsening} and the SOTA heterogeneous graph condensation method, HGCond~\cite{HGCond}.
For the corset methods, 
The Random method randomly picks graphs from the training dataset. Herding selects samples that are closest to the cluster center. K-Center selects the center samples to minimize the largest distance between a sample and its nearest center.
Since the corset method is designed for homogeneous graphs, we develop three variants named Random-HG, Herding-HG, and K-center-HG to support heterogeneous graph reduction. Specifically, we adopt HGNNs to generate learned embeddings and use them as input to Herding-HG and Kcenter-HG.
For the graph coarsening method, we adopt the variation neighborhoods method implemented by~\cite{graph-coarsening} and use the heterogeneous graph as input, this variant named Coarsening-HG.

\subsection{Experimental Settings}
\textbf{Evaluation.}
We first use the aforementioned baselines to obtain condensed graphs and then evaluate them on the test model, we condense the full graph with the condensation ratio $r (0<r<1)$ into a condensed graph.
For HGCond, we choose HeteroSGC as the relay model since it performs the best in HGCond. 
For coreset methods, we adopt SeHGNN (the SOTA HGNN) to obtain intermediate embeddings. 
For the test model, all methods adopt SeHGNN. 
We use the condensed graph to train the SeHGNN model and subsequently employ the trained model for testing on the full graph. The obtained test performance is then compared with that of training on the full graph.
For each dataset and its corresponding condensation ratio, we generate 5 condensed graphs with different seeds and report average test performance and variance.

\textbf{Condensation Ratio}.
In heterogeneous graphs, other-types nodes usually do not contain labels. In order to ensure a uniform condensation ratio across the entire graph, we condense the number of nodes type by type according to the condensation ratio.
For a dataset with a total number of $N$ nodes, the condensed graph contains $rN_{type}$ nodes of each type, and the total number of nodes in the condensed graph is $rN$.
Due to the different characteristics of datasets, different datasets often need to adjust the condensation ratio $r$ to achieve satisfactory performance~\cite{Gcond, HGCond, one-step, SFGC, Eigenbasis-Matching, graph-skeleton},  we follow GCond's condensation approach, which is based on the labeling rate. For example, ACM's labeling rate is 24\%, and we set $\{5\%, 10\%, 20\%\}$ of the labeling rate as the condensed graph. Thus, we finally choose $r = \{1.2\%, 2.4\%, 4.8\%\}$ for ACM. For large-scale dataset AMiner, $r = \{0.05\%,0.2\%,0.8\%\}$. For knowledge graphs MUTAG and AM, $r = \{0.5\%,1.0\%,2.0\%\}$ and $\{0.2\%,0.4\%,0.8\%\}$, respectively.

\textbf{Hyperparameter settings.}
Since our method performs graph condensation in the pre-processing stage, there are no hyperparameter in the relay model training stage.
In the pre-processing stage, the hyperparameters include the hop numbers $K$ to generate meta-paths. We set $K=\{3, 4, 5, 2, 1, 1, 2\}$ for ACM, DBLP, IMDB, Freebase, MUTAG, AM and AMiner, respectively.
During the model evaluation stage, the hyperparameters include the learning rate $lr$, the dimension of hidden layers $D$, and the dropout rate $drop$.
For all datasets, $lr$ is set to 0.001 and $drop$ is set to 0.5. $D$ is set to 128 for middle-scale datasets, and 512 for large-scale dataset.

\textbf{Experimental environment.} The experiments are conducted on a machine with Intel(R) Xeon(R) Gold 5120 CPU @ 2.20GHz and a single TITAN RTX GPU with 24GB GPU memory. The operating system of the machine is Ubuntu 16.04. As for software versions, we use Python 3.7, Pytorch 1.12.1, and CUDA 10.1.

\subsection{Comparison with Baselines on HGB Benchmark}
In this subsection, we conduct two experiments. In the first experiment, we limit the condensation ratio to observe the performance. In the second experiment, we further increase $r$ to observe when the accuracy approaches the whole graph.

\textbf{Satisfactory performance at low condensation ratio.}
We present the evaluation results of our node classification predictions on four middle-scale heterogeneous graph datasets.
As shown in Table~\ref{exp1}, FreeHGC outperforms all baseline methods across most condensation ratios, except DBLP and IMDB at a condensation ratio of $r=1.2\%$, where its accuracy is lower than HGCond. The reason is that FreeHGC focuses on data selection and synthesis of the original graph, and may not be able to capture key graph structure information when the condensation ratio is too low.
The prediction results showcase the effectiveness of FreeHGC.
It is worth noting that only 4.8\% condensation ratio is needed for FreeHGC to reach 98.76\%, 98.75\%, 92.70\%, 96.56\% of the whole dataset accuracy of ACM, DBLP, IMDB, and Freebase datasets.

\begin{figure}[t]
    \centering
    \begin{subfigure}{0.45\linewidth}
        \includegraphics[width=\linewidth]{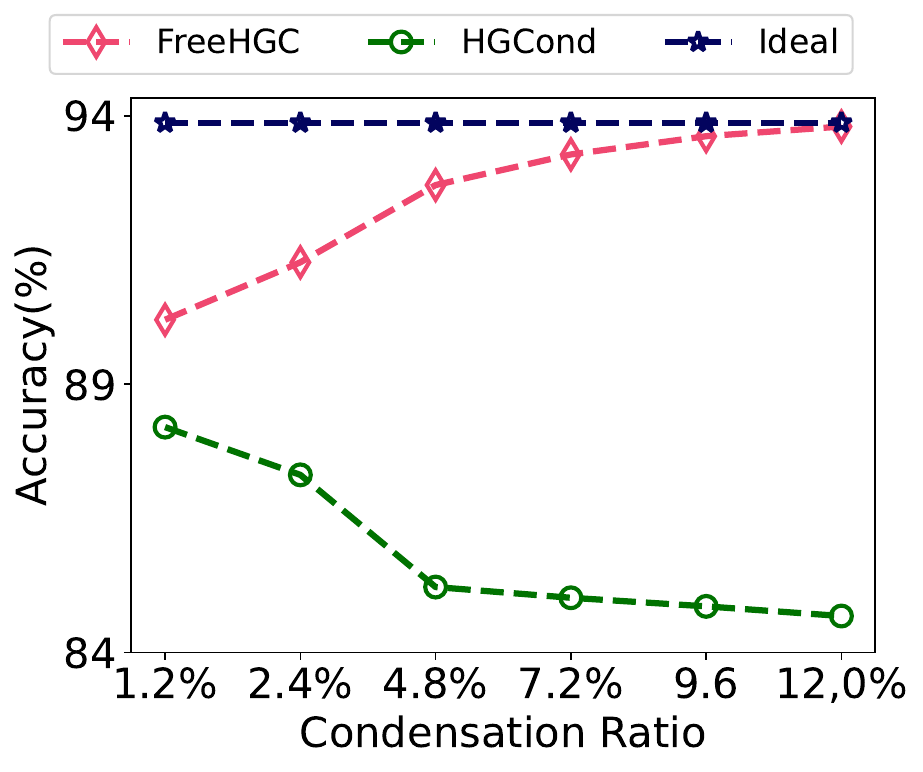}
    \end{subfigure}
    \begin{subfigure}{0.45\linewidth}
        \includegraphics[width=\linewidth]{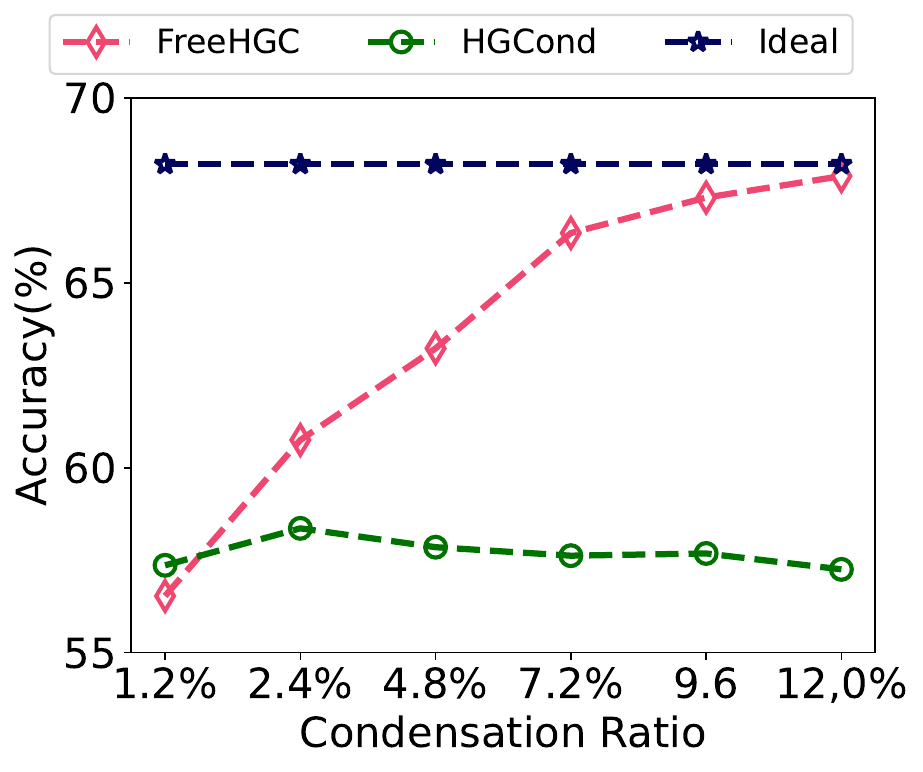}
    \end{subfigure}
    \caption{Performance at different condensation ratio. Left: ACM dataset. Left: IMDB dataset.}
\label{exp2}
\end{figure}

\noindent \textbf{Satisfactory condensation ratio at high performance.}
Since our method requires no training and is more efficient than traditional graph condensation methods, the performance under more condensation ratio can be tested.
We select ACM and IMDB datasets and increase $r$ from 1.2\% to 12\% to observe the prediction accuracy. As shown in Figure~\ref{exp2}, the accuracy rate continues to increase as $r$ increases.
When $r=12\%$, the accuracy of ACM is 93.87\%, which is 99.93\% of the whole dataset accuracy.
The performance of IMDB is 67.89\%, which is 99.53\% of the whole dataset accuracy.
The test accuracy is almost equal to the performance of the original graph.
This illustrates the effectiveness of our method at flexible condensation ratio setting.

\begin{table}[!t]
\renewcommand\arraystretch{1.2}
\caption{\textbf{Experiment results of node classification prediction tasks on knowledge graphs.}}
\setlength{\tabcolsep}{1mm}
\centering
\resizebox{0.5\textwidth}{!}{
\begin{tabular}{c|*{3}{c}|*{3}{c}}
\hline
&\multicolumn{3}{c}{\textbf{MUTAG} (Whole ACC: 73.28)}&\multicolumn{3}{c}{\textbf{AM} (Whole ACC: 88.52)}\\
\hline
\textbf{Method}&r = 0.5\%&r = 1.0\%&r = 2.0\%&
r = 0.2\%&r = 0.4\%&r = 0.8\%\\
\hline
Herding-HG&$50.28\pm0.8$&$51.56\pm2.5$&$52.35\pm3.4$&
$57.62\pm1.4$&$58.33\pm0.9$&$59.05\pm2.7$\\
\hline
GCond&$60.23\pm3.3$&$60.61\pm2.5$&$60.33\pm3.6$&
$68.23\pm4.2$&$68.15\pm2.8$&$68.19\pm4.5$\\
\hline
HGCond&$66.01\pm1.6$&$66.58\pm0.8$&$66.54\pm2.4$&
$78.56\pm0.6$&$78.33\pm1.2$&$78.15\pm3.7$\\
\hline
FreeHGC&\pmb{$66.92\pm0.5$}&\pmb{$68.05\pm1.1$}&\pmb{$69.21\pm0.8$}&
\pmb{$79.61\pm0.4$}&\pmb{$81.35\pm0.3$}&\pmb{$82.55\pm1.6$}\\
\hline
\end{tabular}}
\vspace{-0.3cm}
\label{exp:kg}
\end{table}


\begin{figure}[t]
    \centering
    \begin{subfigure}{0.32\linewidth}
        \includegraphics[width=\linewidth]{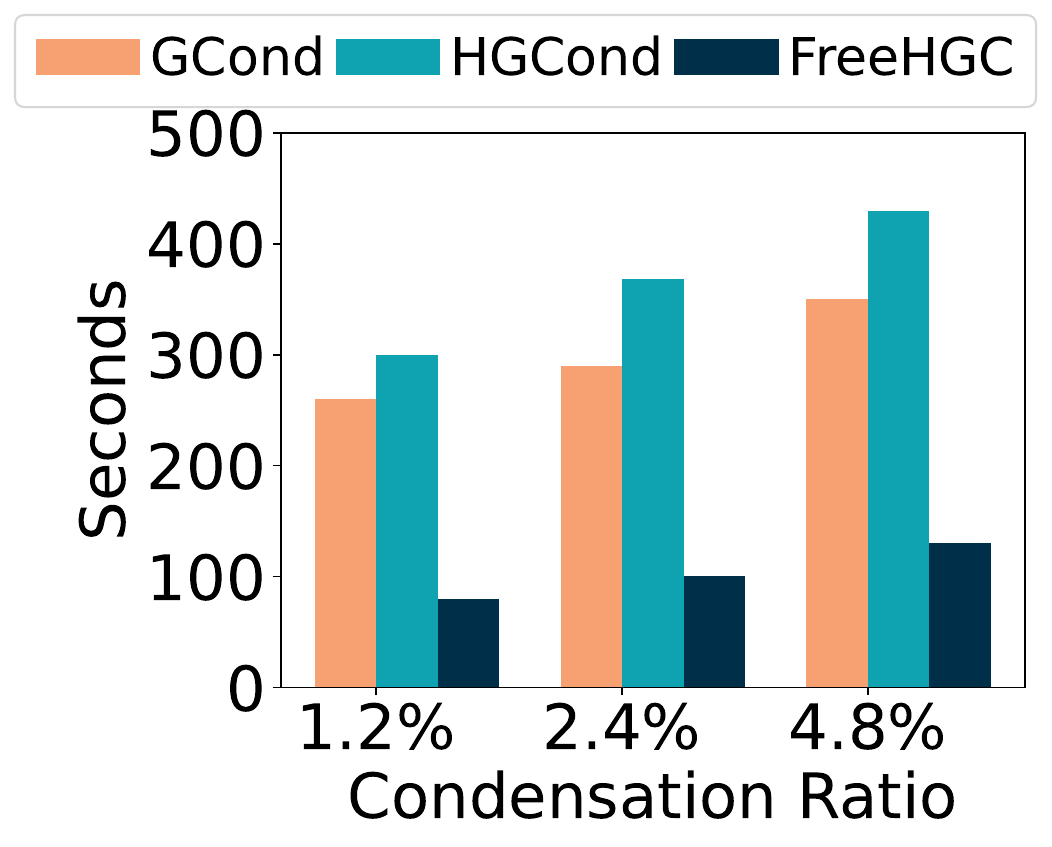}
    \end{subfigure}
    \begin{subfigure}{0.32\linewidth}
        \includegraphics[width=\linewidth]{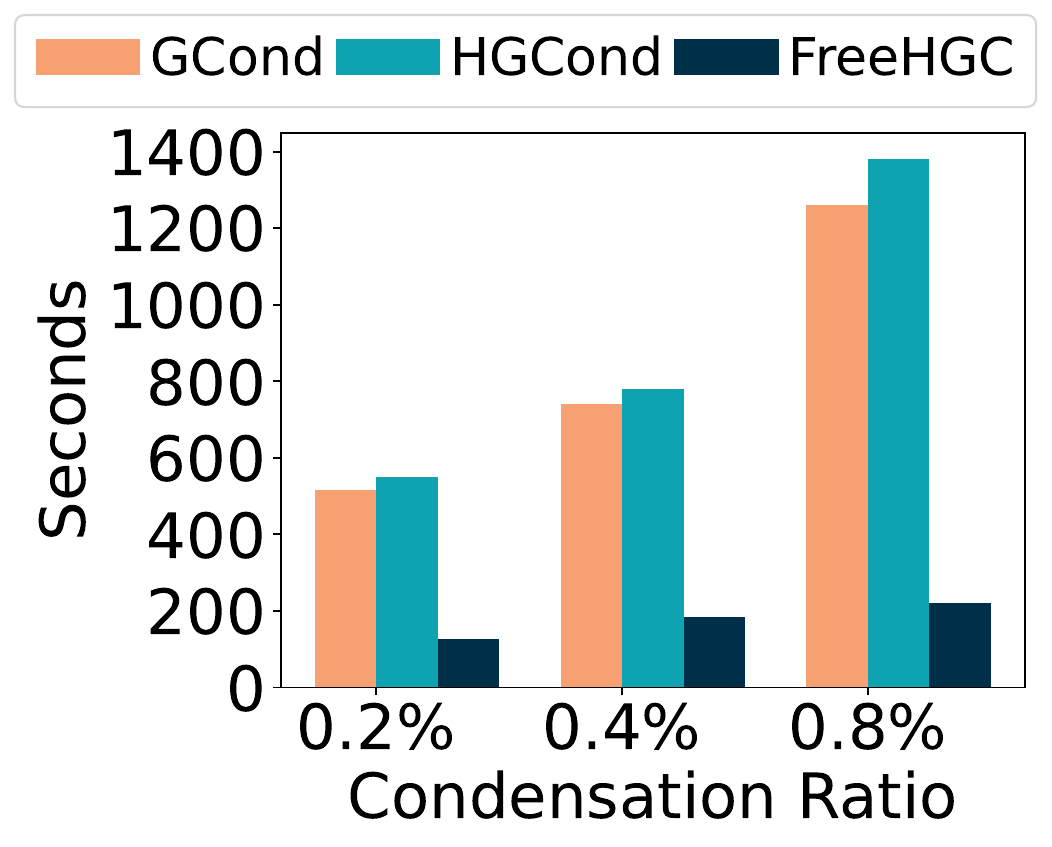}
    \end{subfigure}
    \begin{subfigure}{0.32\linewidth}
        \includegraphics[width=\linewidth]{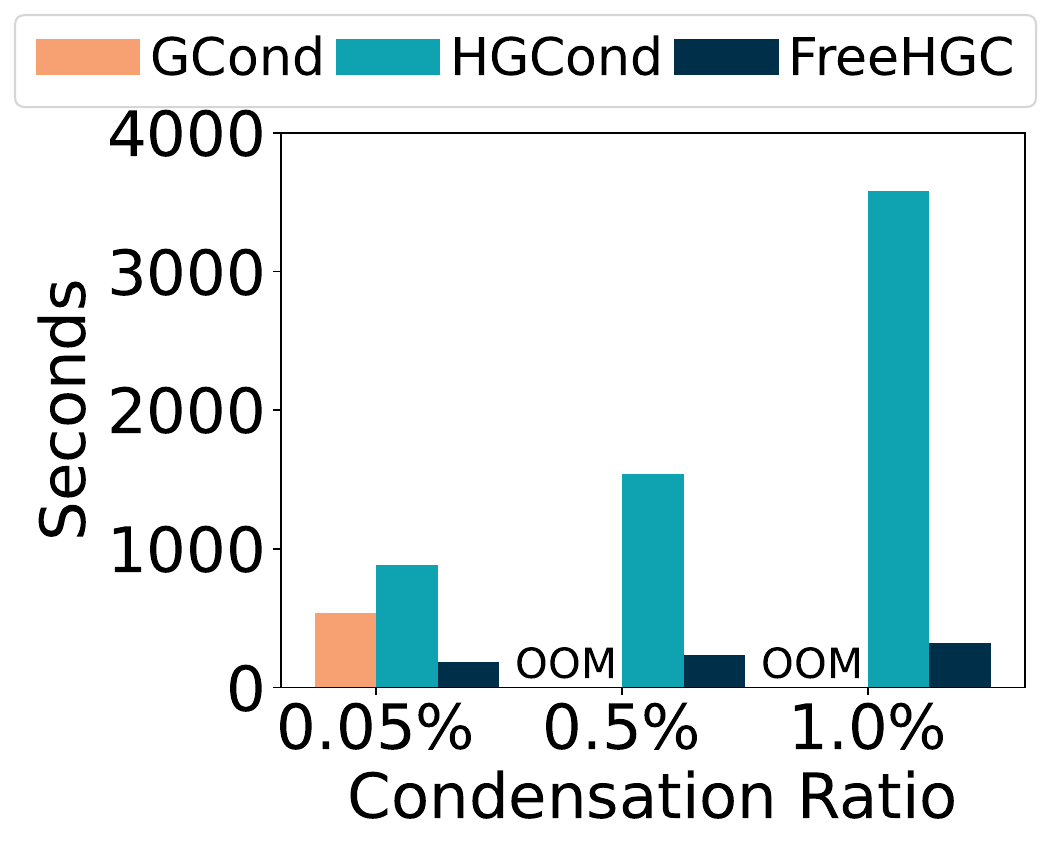}
    \end{subfigure}
    \caption{Comparison of time cost with other methods. Left: Freebase dataset. Middle: AM dataset. Right: AMiner dataset.}
\label{exp:efficiency}
\end{figure}

\begin{table}[t]
\renewcommand\arraystretch{1.3}
\caption{\textbf{Experiment results on the large-scale dataset AMiner compared with other baselines.}}
\centering
\resizebox{0.48\textwidth}{!}{
\begin{tabular}{c|*{3}{c}|{c}}%
\toprule%
Methods&r=0.05\%&r=0.2\%&r=0.8\%&Whole acc\\
\hline
Herding-HG&$68.23\pm0.31$&$72.92\pm0.24$&$83.90\pm0.28$&\multirow{4}{*}{$\pmb{94.30\pm0.24}$}\\
GCond&$63.25\pm0.57$&OOM&OOM\\
HGCond&${87.53\pm1.15}$&${87.50\pm0.89}$&${87.02\pm1.39}$\\
FreeHGC&$\pmb{90.18\pm0.35}$&$\pmb{90.53\pm0.42}$&$\pmb{91.86\pm0.17}$\\
\hline
\end{tabular}}
\label{exp:scala}
\vspace{-0.2cm}
\end{table}

\subsection{Generalization Ability of FreeHGC across HGNNs}
We evaluate the generalization ability of the proposed FreeHGC.
Considering that Herding-HG performs better than other coreset methods, we select Herding-HG for experimental comparison. We also select GCond and HGCond for comparison.
Concretely, we test the node classification performance of our condensed graph with four different commonly used HGNN models: HGB, HGT, HAN, and SeHGNN. The first two are meta-paths-free methods, and the latter two are meta-paths-based methods.
We calculate the average accuracy of the condensed graph and the whole graph of different models to compare the generalization ability of different methods.

Table~\ref{exp3} shows that the proposed FreeHGC achieves outstanding performance over all tested HGNN models, reflecting its excellent generalization ability.
For example, the DBLP dataset enriched by FreeHGC has an average accuracy of 91.77\% across all models, accounting for 97.83\% of the accuracy of the whole graph accuracy.
This is because our method focuses on data selection and is not affected by different relay model designs, making it well applicable to different HGNN models.

\begin{table*}[t]
\renewcommand\arraystretch{1.2}
\caption{\textbf{Comparison between condensed graphs and original graphs. TH: Training HGB (100 epochs). TS: Training SeHGNN (100 epochs).}}
\setlength{\tabcolsep}{1mm}
\centering
\resizebox{0.95\textwidth}{!}{
\begin{tabular}{{c}{c}{c}{c}|{c}{c}{c}|{c}{c}{c}|{c}{c}{c}|{c}{c}{c}}%
\toprule%
&\multicolumn{3}{c}{ACM, r = 2.4\%}&\multicolumn{3}{c}{DBLP, r = 2.4\%}&\multicolumn{3}{c}{IMDB, r = 2.4\%}&\multicolumn{3}{c}{Freebase, r = 2.4\%}&\multicolumn{3}{c}{AMiner, r = 0.2\%}\\
\cline{2-16}
&Whole&HGCond&FreeHGC&Whole&HGCond&FreeHGC&Whole&HGCond&FreeHGC&Whole&HGCond&FreeHGC&Whole&HGCond&FreeHGC\\
\hline
Accuracy&$93.87$&$87.31$&$91.27$&$95.24$&$93.01$&$93.21$&$68.21$&$58.36$&$60.75$&$63.41$&$54.34$&$60.55$&$94.30$&$87.50$&$90.53$\\
Storage&66.2MB&1.75MB&1.6MB&251MB&6.4MB&5.7MB&417.9MB&20.1MB&18.5MB&726.9MB&71.3MB&30.6MB&2770.6MB&170.2MB&30.5MB\\
TH&10.6s&3.8s&3.1s&14.6s&5.1s&4.3s&11.5s&3.9s&3.2s&224.6s&16.8s&14.7s&OOM&68.7s&62.5s\\
TS&4.0s&1.8s&1.2s&6.2s&2.6s&2.3s&5.8s&2.2s&1.6s&96.6s&8.3s&6.7s
&186.1s&21.6s&17.4s\\
\hline
\end{tabular}}
\label{exp5}
\end{table*}

\begin{table*}[!t]
\renewcommand\arraystretch{1.2}
\caption{\textbf{Ablation study to verify the effectiveness of FreeHGC.}}
\setlength{\tabcolsep}{1.2mm}
\centering
\resizebox{0.95\textwidth}{!}{
\begin{tabular}{c|c|*{6}{c}|*{6}{c}|*{6}{c}}
\hline
&&\multicolumn{6}{c}{\textbf{ACM}}&\multicolumn{6}{c}{\textbf{DBLP}}&\multicolumn{6}{c}{\textbf{AMiner}}\\
\hline
&\textbf{Method}&r = 1.2\%&$\Delta$&r = 2.4\%&$\Delta$&r = 4.8\%&$\Delta$
&r = 1.2\%&$\Delta$&r = 2.4\%&$\Delta$&r = 4.8\%&$\Delta$
&r = 0.05\%&$\Delta$&r = 0.2\%&$\Delta$&r = 0.8\%&$\Delta$\\
\hline
\multirow{3}{*}{\makecell[c]{\textbf{Condense} \\\textbf{Target-type}}}
&
Variant\#1&$88.1$&-1.9&$88.5$&-2.8&$89.1$&-3.6&
$87.6$&-3.8&$88.3$&-4.9&$89.5$&-4.6&
85.3&-4.9&86.0&-4.5&87.5&-4.4\\
&
Variant\#2&$86.4$&-3.6&$86.7$&-4.6&$87.2$&-5.5&
$88.3$&-3.1&$91.2$&-2.0&$92.7$&-1.4&
86.4&-3.8&86.9&-3.6&88.5&-3.6\\
&
Variant\#3&$78.3$&-11.7&$79.1$&-12.2&$82.6$&-10.1&
$79.6$&-11.8&$81.1$&-12.1&$83.5$&-10.6&
80.4&-9.8&81.7&-8.8&82.6&-9.3\\
\hline
\textbf{Baseline}&\textbf{FreeHGC}&\textbf{90.0}&&\textbf{91.3}&&\textbf{92.7}&&\textbf{91.4}&&\textbf{93.2}&&\textbf{94.1}&&\textbf{90.2}&&\textbf{90.5}&&\textbf{91.9}\\
\hline
\multirow{3}{*}{\makecell[c]{\textbf{Condense} \\\textbf{Other-types}}}
&
Variant\#4&$85.9$&-4.1&$86.1$&-5.2&$86.9$&-5.8&
$87.2$&-4.2&$88.4$&-4.8&$88.9$&-5.2&
85.9&-4.3&86.6&-3.9&88.0&-3.9\\
&
Variant\#5&$87.5$&-2.5&$88.2$&-3.1&$90.5$&-2.2&
$81.3$&-10.1&$83.4$&-9.8&$84.7$&-9.4&
81.7&-8.5&82.3&-8.2&83.5&-8.4\\
&
Variant\#6&$80.8$&-9.2&$82.9$&-8.4&$83.1$&-9.6&
$73.6$&-17.8&$76.8$&-16.4&$79.2$&-14.9&
74.2&-16.0&75.1&-15.4&78.6&-13.3\\
\hline
\end{tabular}}
\vspace{-0.3cm}
\label{exp6}
\end{table*}

\begin{figure}[t]
    \centering
    \begin{subfigure}{0.45\linewidth}
        \includegraphics[width=\linewidth]{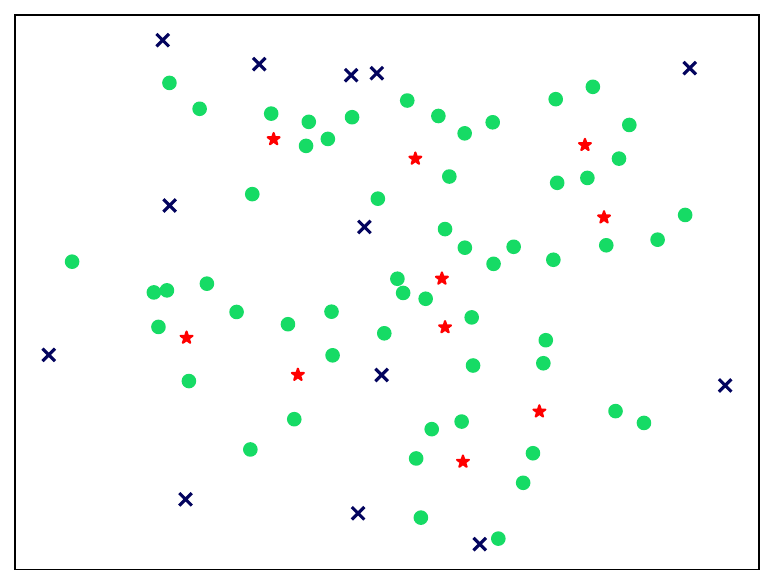}
    \end{subfigure}
    \begin{subfigure}{0.45\linewidth}
        \includegraphics[width=\linewidth]{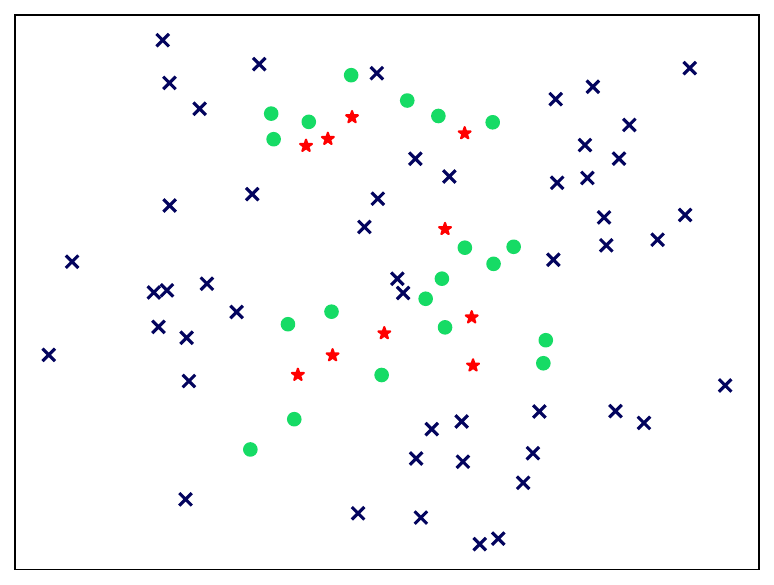}
    \end{subfigure}
    \caption{Visualization of target-type nodes and captured nodes selected by (left) FreeHGC and (right) Herding.}
\vspace{-0.5cm}
\label{t-SNE}
\end{figure}

\subsection{Comparison with Baselines on Knowledge Graphs}
We also evaluate FreeHGC on two widely used knowledge graphs in Resource Description Framework (RDF) format~\cite{rdf}: MUTAG and AM, as shown in Table~\ref{tab:Datasets}. As depicted in Table~\ref{exp:kg}, our proposed FreeHGC outperforms other baselines on both datasets. This shows the effectiveness of our approach on knowledge graphs with more relations.

\subsection{Efficiency Analysis}
To further demonstrate the superior efficiency of FreeHGC, we compared it with GCond and HGCond on Freebase, MUTAG and the large-scale dataset AMiner. Figure~\ref{exp:efficiency} reports the time cost of different methods when they perform best. The condensation time of HGCond is higher than that of GCond because it requires additional cluster information and an OPS-based parameter exploration strategy. The time consumption is particularly serious on the large-scale graph AMiner, it takes about 1 hour to condense 1\% of AMiner.
In contrast, FreeHGC has the lowest time cost because it does not require training. For the above three datasets, FreeHGC has at most $4.16\times$ and $4.67\times$,  $5.73\times$ and $6.27\times$,  $3.12\times$ and $11.19\times$ faster than GCond and HGCond, respectively.

\subsection{Scalability of FreeHGC}
To demonstrate the scalability of FreeHGC, we conducted a comparison with Herding-HG, GCond, and HGCond on large-scale heterogeneous graph dataset AMiner~\cite{AMiner}. The condensation ratio is {0.05\%, 0.2\%, 0.8\%}. 

Figure~\ref{exp:scala} shows that FreeHGC performs best at different condensation ratios, and the accuracy gradually increases. HGCond is suboptimal and cannot further improve the accuracy due to the condensation ratio limitations.
For GCond, when the condensation ratio is larger than 0.2\%, it will be out of memory because it uses a more dense connection format.

\subsection{Analysis on Condensed Data}
In Table~\ref{exp5}, we compare the test accuracy, storage cost, and model training time between condensed graphs and original graphs. We selected HGB and SeHGNN models for training.
For the storage cost, FreeHGC reduces the storage by 97.6\%, 97.7\%, 95.7\%, 95.6\% and 98.9\% for five datasets.
For the training time comparison, the condensed graph achieves acceleration on both HGB and SeHGNN.
For example, on the SeHGNN model, the training time on five datasets is only 30.0\%, 37.1\%, 27.6\%, 6.59\%, and 9.3\% of that on the whole graph.
We also compared FreeHGC with HGCond and found that HGCond has higher memory usage due to its reliance on traditional graph condensation methods, which generate a denser small graph that differs from the original graph. As a result, HGCond required more training time than FreeHGC.

\subsection{Method Interpretability}
This section explains why FreeHGC can achieve better performance with the data selection criterion F(S). To describe the data distribution, we randomly select 80 nodes from ACM dataset and select 10 target-type nodes (red star) using FreeHGC and the advanced data selection
method Herding, as shown in Figure~\ref{t-SNE}. We then mark all nodes that can be captured within 3 hops (green circle, including activated other-types and target-type nodes), the black crosses refer to the un-captured nodes. We visualize them using t-SNE~\cite{t-sne}.

For $R(S)$, FreeHGC activates more nodes than Herding, which means more receptive fields are included.
For $1-J(S)$, the nodes captured by FreeHGC are scattered in different areas of the entire dataset, while Herding concentrates them in specific areas.
The above two observations further explain that FreeHGC's data selection criterion is more effective in capturing the graph structure information of other-types nodes compared to other data selection methods.

\subsection{Ablation Study}
To conduct a comprehensive study of FreeHGC, we performed ablation studies to analyze the effectiveness of our two key parts: condense the target-type and condense other-types.
We conduct experiments on the ACM, DBLP and large-scale dataset AMiner. We also use the $\Delta$ metric to evaluate the differences between the variants and the baseline.

\textbf{Ablation analysis of condensed target-type nodes.}
We set up different variants for analysis: Variant\#1 does not use the receptive field maximization function. Variant\#2 does not use the meta-paths similarity minimization function. Variant\#3 uses the advanced data selection method Herding for target types. FreeHGC is used as our baseline.

On the ACM dataset, the contribution of Variant\#1 is greater than Variant\#2. The reason is that ACM has more relations than DBLP and AMiner, which leads to more meta-paths. The meta-path similarity minimization function can help ACM better capture the diversity of nodes.
On the DBLP and AMiner datasets, the contribution of Variant\#2 is greater than Variant\#1. The receptive field maximization function can help activate more nodes. Besides, our proposed criterion shows effectiveness compared to Herding (Variant\#3).

\textbf{Ablation analysis of condensed other-types nodes.}
Condense other-types including two components: (a) neighbor influence maximization and (b) information loss minimization. 
We set up different variants for analysis: 
For DBLP and AMiner, Variant\#4 uses (a), and Variant\#5 uses (b) to condense father-type, and uses Herding to condense leaf-type.
For ACM, Variant\#4 uses (a), and Variant\#5 uses (b) to condense the "author" type and uses Herding for other leaf-types.
Variant\#6 uses Herding for both father-type and leaf-types.


On the ACM dataset, Variant\#4 performs worse than Variant\#5, indicating that (b) for leaf-type nodes provides richer graph structure and attribute information. On the DBLP and AMiner datasets, Variant\#4 outperforms Variant\#5, which proves that (a) for father-type nodes effectively captures graph structure. Variant\#6 shows our method surpasses Herding.

\section{Conclusion}
This paper presents FreeHGC, a new training-free heterogeneous graph condensation method.
The goal is to select and synthesize high-quality nodes from the original large graph and condense it into a small graph without training. Therefore, we propose FreeHGC to facilitate both efficient and high-quality generation of condensed graphs.
Experimental results on a node classification task demonstrate that FreeHGC can significantly reduce graph size while maintaining satisfactory performance, and has the advantage of flexible condensation ratio.
In addition, experimental results also illustrate that our method has good generalization and scalability capabilities.

\section*{ACKNOWLEDGMENT}
This work is supported by National Natural Science Foundation of China (U23B2048, U22B2037, 92470121, 62402016), research grant No. SH-2024JK29, Alibaba Cloud, and High-performance Computing Platform of Peking University.

\bibliographystyle{IEEEtran}
\bibliography{HGNNbibtex.bib}

\end{document}